\documentclass{article}

\PassOptionsToPackage{numbers,compress}{natbib}
\usepackage[preprint]{neurips_2026}

\usepackage[utf8]{inputenc}
\usepackage[T1]{fontenc}
\usepackage{hyperref}
\hypersetup{hidelinks,pdftitle={Decision System Engineering: Managing Bias and Noise},pdfauthor={Mikhail Arbuzov}}
\usepackage{url}
\usepackage{booktabs}
\usepackage{amsfonts}
\usepackage{amssymb}
\usepackage{amsmath}
\usepackage{nicefrac}
\usepackage{microtype}
\usepackage{xcolor}
\usepackage{graphicx}

\newcommand{\lgt}{\operatorname{logit}}
\newcommand{\Var}{\operatorname{Var}}
\newcommand{\E}{\mathbb{E}}
\newcommand{\gate}[1]{\textsc{#1}}

\title{Decision System Engineering:\\Managing Bias and Noise}

\author{Mikhail Arbuzov\\Independent Researcher}

\begin{document}

\maketitle

\begin{abstract}
When a journal, funding agency, or employer makes a poor choice, blaming the judges leaves open
whether their replacements would make the same mistake. Decision System Engineering models the
procedure those judges inherit. Each score reflects the quality of what is being judged, lasting
preferences that judges may share, and chance variation from one judgment to the next. A
decision circuit links any number of stages by specifying who judges at each one, how their
scores are combined, and who advances or gets to judge next. Following those scores through the
circuit shows which errors cancel and which shape the outcome. The model offers explanations for
31 patterns reported across seven fields. Selecting the highest of several noisy bids favors an
overestimate, producing the winner's curse. Two review panels can apply the same cutoff to the
same papers and accept different ones. On a team that hires its own replacements, giving two
groups equal seats on the hiring panel removes a chance majority's hiring advantage. Once we
understand why a decision system keeps making the same mistake, we can redesign it to prevent
that mistake.
\end{abstract}

\section{Introduction}
\label{sec:intro}

Decisions that matter are made by systems of fallible evaluators: a hiring pipeline of r\'esum\'e
screen, interview panels and a final interview; a grant panel; a conference's program committee.
Machine learning runs the same kind of system on itself when it selects models on validation
scores, ranks them on leaderboards and scores them with juries of language-model judges. These
systems fail in well-documented ways, each under its own name in its own literature: the winner's
curse of auctions~\citep{thaler1988}, the committee of many that judges like one~\citep{hogarth1978},
the disfavored group whose admitted members outperform~\citep{becker1957}, the team that hires in
its own image~\citep{kanter1977}, the two program committees that accept different
papers~\citep{cortes2021}, the selected model that disappoints on fresh
data~\citep{cawley2010overfitting}. Each literature has its remedies, and most of them act on one
stage.

Whether such a remedy works, though, depends on what the evaluators share and on the stages around
it. A grant panel that grows from three members to twelve to average out their quirks keeps almost
all of the bias they share if the twelve were trained alike, and a leaderboard that adds judges
lowers its winner's overstatement only as far as the judges' tastes differ (\S\ref{sec:average},
\S\ref{sec:jury}). On a team that hires its own replacements, lowering the preference for one's own type
shortens the life of a majority. A larger applicant pool lets a smaller preference reinforce a
$7$--$5$ majority (\S\ref{sec:homophily}). A designer needs to see how a change at one gate travels through
the others, and so which change reaches a failure and which leaves it in place.

Earlier models compute parts of this. Generalizability theory's decision
studies~\citep{brennan2001generalizability} and a decision study of multiple
mini-interviews~\citep{eva2004relationship} price a change of raters or interviews at one stage;
multistage selection computes a design's quality and adverse impact across its
stages~\citep{decorte2006}; and organizational economics wires fallible screens in series and in
parallel~\citep{sah1986,christensen2010}, with errors independent across screens. None follows an
error from its seat through composed gates and feedback to the outcome.

In Decision System Engineering (DSE), the object of design is the arrangement of fallible
evaluators: the rules, the wiring of the stages and who sits with whom at each gate. Designing
arrangements of fallible agents is an old aim~\citep{sah1986}; this paper adds a first model of
their errors, the \emph{decision circuit}, and the model needs little. Its evaluator sees true quality through a
bias that stays with them and through noise drawn afresh at each judgment, and the bias has a part
common to all evaluators, a part shared with those who sit beside them and a part of their own
(\S\ref{sec:circuit}). Its gates are of four kinds, each with one law (\S\ref{sec:gates}), and they
may be wired in any way, including gates staffed by the system's own output. That is enough to write
down hiring pipelines, review panels and machine learning's evaluation pipelines alike, to state a
sufficient mechanism for each of $31$ reported patterns across seven literatures (\S\ref{sec:reach}), and,
for stated parameters, to compute what a change of rule, seat or stage will do before it is made
(\S\ref{sec:compose}). Because the phenomena need so little of the evaluator, an agent simulation
of people that reproduces them shows little; \S\ref{sec:ml} says what would test one, and how
to measure what model judges share.

\section{The decision circuit: what travels with a judge}
\label{sec:circuit}

A decision circuit has three parts: evaluators, gates and the wiring between them. Writing a system
as a circuit asks its designer to say, at every gate, who recurs and what the judges there share,
because in the model those two facts decide which errors reach the outcome.

\paragraph{The evaluator.} An evaluator $i$ judges a candidate whose true success probability is
$q$. It sees $q$ through a bias on the candidate's observable features $x$ and an occasion noise,
in log-odds:
\begin{equation}
\lgt(\hat q_i) \;=\; \lgt(q) \;+\; \beta_i^\top x \;+\; \varepsilon_i .
\label{eq:model}
\end{equation}
The evaluator has no memory, language or strategy. It is the lens-model judge with no nonlinear
component~\citep{karelaia2008}, except that it sees true quality itself rather than cues to it. Its
bias $\beta_i^\top x$ is \emph{persistent}: the evaluator adds it to every judgment of a candidate
with features $x$. Its \emph{occasion noise} $\varepsilon_i$, of scale $\sigma_\varepsilon$, is
drawn afresh for each judgment. Unless stated otherwise, $\lgt(q)\sim\mathcal N(0,1)$, the noise is
Gaussian, and $x$ is one binary feature, so the bias is a single number.

\paragraph{The parts of the bias.} Across evaluators the bias splits as $\beta_i=\bar\beta+\delta_i$.
The \emph{common part} $\bar\beta$ is held by every evaluator. The varying part $\delta_i$, of scale
$\sigma_\beta$, splits again among the evaluators who sit together at a gate:
$\delta_i=\gamma+\eta_i$, where the \emph{panel part} $\gamma$ is held by all of them and the
\emph{own part} $\eta_i$ by each alone. The panel part's share of the varying part's variance is the
\emph{bias correlation} $\rho_\beta$, the correlation between two members' varying parts. It belongs
to who sits together rather than to any one person. Evaluators trained alike, or language models
trained on the same data, share a large panel part. \emph{What the judges share}, the common part
and the panel part together, is what no average at that gate removes. We call everything that stays
with an evaluator its bias and keep \emph{noise} for occasion noise; \citet{kahneman2021} call
$\bar\beta$ bias and $\delta_i$ level and stable pattern noise.

\begin{figure}[t]
\centering
\includegraphics[width=\linewidth]{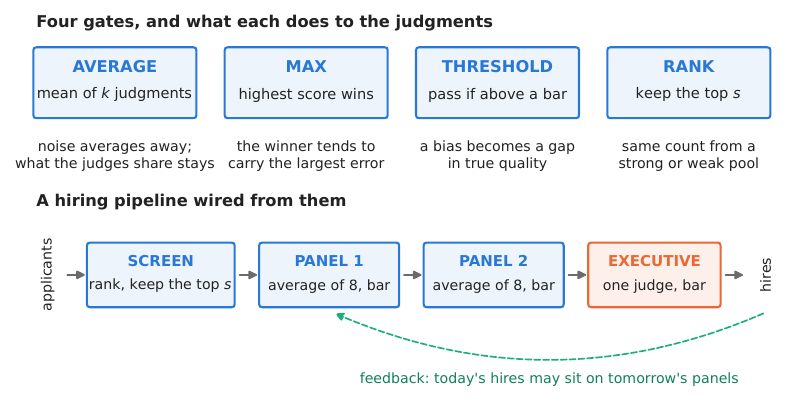}
\caption{\textbf{A decision circuit.} Top: the four gates and what each does to the judgments
it receives (\S\ref{sec:gates}). Bottom: a hiring pipeline wired from them. A screen keeps the top $s$
applicants, two panels of eight average their judgments and apply a bar, and one executive makes
the last cut. The dashed arrow is feedback: today's hires may staff tomorrow's panels.}
\label{fig:circuit}
\end{figure}

\paragraph{Gates and wiring.} A gate collects its evaluators' judgments and applies one of four
rules. \gate{average} pools the $k$ judgments of each candidate into their mean. \gate{max} passes
the highest judgment of a set: several evaluators' judgments of one item, as in an auction, or one
evaluator's judgments of many items, as in model selection. \gate{threshold} passes every candidate
whose judgment clears a bar $\tau$. \gate{rank} passes the top $s$ of a batch. Gates
are wired into a directed graph through which candidates flow, and those one gate passes become the
pool the next gate judges (Figure~\ref{fig:circuit}). When who staffs a gate depends on what the
circuit passed earlier, as when a team interviews its own replacements, the circuit has
\emph{feedback}. The \emph{arrangement} is the graph, the rule at each gate and who sits at it. A
circuit's outcome follows from stated parameters, in closed form for one gate and by simulation for
any wiring; Appendix~\ref{app:params} lists every parameter, and the code behind every number will
be released.

\paragraph{What travels with a judge.} Two facts decide which errors reach the outcome: what the
judges share, and who recurs. At a single average only the first matters. What the judges share
passes through, the rest averages away, and a noise they shared would act exactly like a shared
bias. Who recurs matters once the same evaluator judges again, because an error that stays with them
is then added again. An interviewer who sees all of an employer's applicants tilts every hire the
same way; an evaluator who sits at two stages carries one bias into both; a team that hires its own
replacements passes its members' biases on to the panel that hires next. Recurrence can also take an
error out of play. In a leaderboard where the same
judges score every model, a bias that raises every model's score by the same amount cannot change
the winner; only a bias toward particular models can.

\section{Four gates, one law each}
\label{sec:gates}

Each kind of gate does one thing to the judgments it receives. Averaging removes independent noise
and stops at what the judges share. Taking the highest picks, among items equally good, the one
carrying the largest error. A bar turns a bias into a gap in true quality among those just admitted,
shrunk by how reliable the scores are. And ranking ignores a shift of the whole pool, which a bar
fixed in advance passes on to the number admitted. Each law is known in its own literature
(Appendix~\ref{app:related}). The circuit adds that one evaluator produces all four, so that they
compose.

\subsection{\gate{average}: noise averages away, what the judges share does not}
\label{sec:average}

Let $k$ evaluators average their judgments of a candidate. The common part $\bar\beta$ passes
through untouched. The varying part and the noise both average, but each only down to its
correlation:
\begin{equation}
\Var(\bar\varepsilon) = \sigma_\varepsilon^2\!\left[\rho_\varepsilon +
\frac{1-\rho_\varepsilon}{k}\right],
\qquad
\Var(\bar\delta) = \sigma_\beta^2\!\left[\rho_\beta + \frac{1-\rho_\beta}{k}\right].
\label{eq:avg}
\end{equation}
This is the variance of an average of equicorrelated terms~\citep[\S15.2]{hastie2009elements}. With
uncorrelated errors the bracket falls as $1/k$ and the error averages away, as in the wisdom of
crowds~\citep{galton1907}. With $\rho>0$ the bracket cannot fall below $\rho$ however large $k$
grows: correlated experts add little~\citep{hogarth1978}, and a crowd's error falls toward the bias
common to its population~\citep{vul2008}. We set $\rho_\varepsilon=0$ throughout, so averaging
removes the noise and stops at what the judges share.

The floor comes quickly. A committee whose biases are correlated at $\rho_\beta$ does the
bias-removing work of $k_{\mathrm{eff}}=k/[1+(k-1)\rho_\beta]$ independent members. Twelve
evaluators trained alike enough that $\rho_\beta=0.9$ remove bias about as well as $1.1$ of them,
while their averaged noise falls to $0.29$ of one member's (Appendix~\ref{app:deriv}): the panel
grows more confident and stays almost as biased as one member. A jury of language-model judges
meets the same floor. Juries from different model families agree with human raters better than one
strong judge~\citep{verga2024juries}, but models' errors are correlated~\citep{kim2025correlated}.

\subsection{\gate{max}: the winner tends to carry the largest error}
\label{sec:max}

Let $N$ evaluators judge one item and let the highest judgment win, as bidders do for an item worth
the same to each of them. If they share their bias, the judgments differ only in their noise, so
\begin{equation}
\arg\max_i \; \lgt(\hat q_i) \;=\; \arg\max_i \; \varepsilon_i .
\label{eq:identity}
\end{equation}
Winning and having the largest error are the same event. The winner overestimates by
$\sigma_\varepsilon m_N$ on average, where $m_N$ is the expected maximum of $N$ standard normals,
and the overestimate grows with the number of bidders: the winner's curse~\citep{thaler1988} and the
optimizer's curse~\citep{smith2006optimizer}. The same law turns precision against the bidder who
has it. Among ten bidders, one with a perfect estimate wins only when all nine others err low, about
one contest in five hundred ($2^{-9}$; Appendix~\ref{app:deriv}). Both results hold for
evaluators who neither shade their bids nor shrink their estimates.

\gate{max}'s second form picks the best of $N$ items scored by one evaluator, as model selection and
best-of-$n$ sampling do. The items now differ in true quality, and the \emph{reliability} $\lambda$
is the share of a score's variance that is true quality. The selected item is overestimated by
$\sqrt{V(1-\lambda)}\,m_N$, where $V$ is the variance of the score's whole error, not only of its
occasion noise; equally good items are the case $\lambda=0$ (Appendix~\ref{app:deriv}). This is the
winner's curse of model selection~\citep{cawley2010overfitting} and the regressional part of best-of-$n$
overoptimization~\citep{gao2023scaling}.

\subsection{\gate{threshold}: a bias becomes a gap in true quality}
\label{sec:threshold}

Let a gate admit every candidate whose judgment clears a bar, and let a feature raise the judgment
by $b$ for one group, the favored, and lower it by $b$ for the other. To clear the same bar, a
disfavored candidate needs an unbiased score $2b$ higher than a favored one, but that does not make
them $2b$ better. A candidate whose score sits at the bar owes part of it to noise, and the
disfavored, who start further below the bar, owe more. When the two groups have the same mean true quality, the gap in true
quality between their candidates at the bar is therefore $2b$ shrunk by the reliability:
\begin{equation}
\text{gap at the bar} \;=\; 2b\lambda,
\qquad
\lambda = \frac{\sigma_T^2}{\sigma_T^2+\sigma_\varepsilon^2},
\label{eq:gap}
\end{equation}
where $\sigma_T$ is the spread of true quality. Read backwards, this is Becker's outcome test: if a
group's admits outperform, the gate was biased against them~\citep{becker1957}. In early-stage
investment, among the startups that one male investor shows interest in, those led by women
outperform those led by men~\citep{ewens2020}. In a population whose median success
probability is $5\%$ rather than the default $50\%$, a gate that admits the top $10\%$ with $b=0.7$
leaves the disfavored group's admits with nearly twice the favored group's mean true success
probability, $0.341$ against $0.177$ (Appendix~\ref{app:deriv}). When the groups' mean quality
differs, the gap becomes $2b\lambda+(1-\lambda)(\mu_D-\mu_F)$, so a gap alone does not identify a
bias, and averages over all admits can mislead about the bar~\citep{simoiu2017}.

\subsection{\gate{rank}: a shift of the whole pool does not move the count}
\label{sec:rank}

\gate{rank} is the rule of leaderboards, fixed-size retrieval, fixed cohorts and the self-staffing
team of \S\ref{sec:homophily}. At the top it selects on noise as \gate{max} does, but it uses only
the order of the scores, so a stronger or weaker pool yields the same number of choices, while a bar
set in advance admits more from a strong pool and fewer from a weak one. With $100$ candidates for
$10$ slots and a pool whose mean drifts unseen, \gate{rank} captures $0.96$ of an oracle's true
quality at every drift, while a bar set in advance slides from $0.85$ to $0.64$, mostly through
slots left unfilled (Appendix~\ref{app:deriv}). Where a weak pool should admit fewer, the fixed count
becomes a cost.

\section{Reach: one model, many documented phenomena}
\label{sec:reach}

The four laws, and the compositions \S\ref{sec:compose} computes from them, give a sufficient mechanism for each of $31$
reported patterns across seven literatures (Table~\ref{tab:catalogue}). These span selection,
judgment, auctions, discrimination testing, statistics, peer review and machine-learning evaluation.
The table distinguishes a mechanism that can produce a pattern from a quantitative comparison
with measurements. The peer-review consistency experiments provide the latter.

\begin{table}[t]
\caption{\textbf{The catalogue} (in full: Appendix~\ref{app:catalogue}). Support: \textbf{Mag},
magnitudes compared with measured values; \textbf{Dir}, the documented direction shown at stated
parameters; \textbf{Cal}, one parameter calibrated to one number; \textbf{CF}, derived in closed
form; \textbf{Map}, a sufficient mechanism without an empirical fit. A $\dagger$ marks
explanations that use a persistent or shared evaluator preference to produce the pattern. Error terms: $\varepsilon$, occasion noise (scale $\sigma_\varepsilon$); $\delta$, the
bias's varying part; $\rho_\beta$, bias correlation; $b$, a bias on a group; $\lambda$, reliability;
$h$, own-type preference.}
\label{tab:catalogue}
\centering
\scriptsize
\setlength{\tabcolsep}{3.5pt}
\renewcommand{\arraystretch}{0.8}
\begin{tabular}{@{}p{8.6cm}p{4.05cm}l@{}}
\toprule
\textbf{Phenomenon} [documentation] & \textbf{Circuit:} gate $\cdot$ error term & \\
\midrule
\multicolumn{3}{@{}l}{\textit{\textup{\gate{average}}: noise averages away, what the judges share does not (\S\ref{sec:average})}}\\
Wisdom of crowds \citep{galton1907} & \gate{average} $\cdot$ $\varepsilon$ & Dir \\
Committee paradox: correlated experts add little$^\dagger$ \citep{hogarth1978} & \gate{average} $\cdot$ floor at $\rho_\beta$ & Dir \\
Crowd within: one's second guess is worth less than another person's$^\dagger$ \citep{vul2008} & \gate{average} in one evaluator, $\rho_\beta{=}1$ & Cal \\
Bootstrapping: a model of the judge beats the judge \citep{karelaia2008} & removing $\varepsilon$ & Map \\
Large differences between judges of the same case \citep{kahneman2021} & two evaluators $\cdot$ $\delta$, $\varepsilon$ & Cal \\
Diverse LLM juries beat one judge, within limits$^\dagger$ \citep{verga2024juries} & \gate{average} $\cdot$ $\delta$, $\rho_\beta$ & Map \\
Swapped order cancels a judge's position bias, not its verbosity bias$^\dagger$ \citep{zheng2023judging} & \gate{average} $\cdot$ opposite vs.\ shared bias & CF \\
A silicon crowd as accurate as a human one, sharing a bias$^\dagger$ \citep{schoenegger2024silicon} & \gate{average} $\cdot$ $\varepsilon$, $\rho_\beta$ & Map \\
Hyper-accuracy: a simulated crowd that cannot be wise \citep{aher2023} & \gate{average}, no independent $\varepsilon$ & Map \\
\multicolumn{3}{@{}l}{\textit{\textup{\gate{max}}: the winner tends to carry the largest error (\S\ref{sec:max})}}\\
Winner's curse in auctions, oil leases, takeovers \citep{thaler1988} & \gate{max} $\cdot$ $\varepsilon$ & Dir \\
Optimizer's curse; postdecision disappointment \citep{smith2006optimizer} & \gate{max} or \gate{rank} $\cdot$ $\varepsilon$ & Dir \\
Winner's curse of model selection \citep{cawley2010overfitting} & \gate{max} over items $\cdot$ $\varepsilon$, $\lambda$ & CF \\
Best-of-$n$ reward overoptimization, regressional part \citep{gao2023scaling} & \gate{max} over items $\cdot$ $\varepsilon$ & Map \\
Small units fill both ends of a ranking \citep{wainer2007} & \gate{max} or \gate{rank} $\cdot$ unequal $\sigma_\varepsilon$ & Dir \\
\multicolumn{3}{@{}l}{\textit{\textup{\gate{threshold}}: a bias becomes a gap in true quality (\S\ref{sec:threshold})}}\\
Outcome test: a disfavored group's admits outperform$^\dagger$ \citep{becker1957,ewens2020} & \gate{threshold} $\cdot$ bias $b$ & Dir \\
Infra-marginality: averages mislead about the bar \citep{simoiu2017} & \gate{threshold} $\cdot$ $\lambda$ & CF \\
Berkson's paradox \citep{berkson1946} & \gate{threshold} on a sum & CF \\
A noisier group is over-admitted at a raw-score bar \citep{emelianov2020} & \gate{threshold} $\cdot$ group-specific $\sigma_\varepsilon$ & CF \\
Regression after selection; Peter principle \citep{lazear2004} & \gate{threshold} or \gate{rank} $\cdot$ $\varepsilon$ via $\lambda$ & CF \\
Hierarchy versus polyarchy error trade-off \citep{sah1986} & series vs.\ either-of, $\beta{=}0$ & CF \\
\multicolumn{3}{@{}l}{\textit{Peer review (\S\ref{sec:reach})}}\\
Two committees disagree; more selectivity, more arbitrariness \citep{cortes2021,beygelzimer2023} & \gate{average} $\to$ \gate{threshold} & Mag \\
Committee agreement predicted from a pooled reviewer reliability \citep{bornmann2010} & \gate{average} $\cdot$ $\varepsilon$ & Cal \\
Reviewer agreement can fall in a pool narrowed by prior selection \citep{pier2018,erosheva2021} & upstream \gate{threshold} narrows quality & Dir \\
\multicolumn{3}{@{}l}{\textit{Composition and feedback (\S\ref{sec:compose})}}\\
Compensatory vs.\ multiple hurdle: the design moves quality and adverse impact \citep{decorte2006} & \gate{average} vs.\ \gate{threshold} series & Dir \\
Unstructured interviews can have lower predictive validity$^\dagger$ \citep{schmidt1998} & \gate{average} $\cdot$ judge-specific cue weights & Map \\
CEO authority can increase cross-firm variation in performance$^\dagger$ \citep{adams2005} & one decision maker $\cdot$ expected payoff & Map \\
Leaky pipeline, evaluator part: modest per-stage gaps compound$^\dagger$ \citep{shaw2012} & series of biased \gate{threshold}s $\cdot$ $b$ & Dir \\
Survivors get harder in classifier cascades \citep{viola2001rapid} & \gate{threshold} series lowers $\lambda$ & Map \\
Outcome homogenization across firms$^\dagger$ \citep{kleinberg2021monoculture} & parallel \gate{threshold} $\cdot$ $\rho$ & CF \\
Homosocial reproduction$^\dagger$ \citep{kanter1977} & panel staffed by past hires $\cdot$ own-type $h$ & Map \\
An early majority can become long-lived (lock-in)$^\dagger$ \citep{arthur1989,salganik2008} & panel staffed by past hires $\cdot$ $h$ & Map \\
\bottomrule
\end{tabular}
\end{table}

\paragraph{How strongly the rows are supported.} One row, from peer review, compares
magnitudes with measurements; the others show a direction, re-derive a known result, calibrate one parameter or
give a mechanism with no fit (Appendix~\ref{app:catalogue}).
That the same few parts state every mechanism shows the model is expressive. It does not show that
the model is right, or better than a competitor such as random rater effects composed through
fallible screens~\citep{sah1986}, and for some rows the literature documents another mechanism as
well, such as candidates' own choices in the leaky pipeline~\citep{shaw2012}.

\paragraph{Where evaluator preferences enter.} Among the marked rows are the committee whose
correlated members add little~\citep{hogarth1978}, the jury whose members share preferences for
particular models~\citep{verga2024juries}, and the team that hires in its own
image~\citep{kanter1977}. In each case, changing who sits together or who returns for the next
decision changes how much of a preference reaches the outcome.

\paragraph{The measured case: peer review.} The one comparison with measurements tests the
circuit's structure, and weakly. NeurIPS twice sent the same submissions to two independently
staffed committees~\citep{cortes2021,beygelzimer2023}. As a circuit each committee is an
\gate{average} followed by a \gate{threshold}, so two committees whose averaged scores correlate at
$r$ are two correlated score lists, each cut at its acceptance rate. Given the measured $r$, the
circuit matches how often one committee's accepted papers fall outside the other's list: $48.1\%$
against $49.5\%$ measured in 2014, and about $49\%$ against $50.6\%$ in 2021. Since $r$ absorbs
whatever bias the committees carry, the match tests that structure and no more, and across the
uncertainty in the 2014 correlation the prediction runs from $44\%$ to $52\%$
(Appendix~\ref{app:deriv}). Calibrated instead to one outside number, the circuit overshoots. Taking a pooled
meta-analytic reliability, $0.34$~\citep{bornmann2010}, as an illustrative value for one reviewer,
Eq.~\eqref{eq:avg} predicts that panels of three or four correlate at $0.61$ to $0.67$, above the
$0.55$--$0.59$ measured at NeurIPS; $0.61$ lies within the $0.60$--$0.66$ of an NSF
re-review~\citep{cole1981}.

\section{Composed, the gates compute a redesign}
\label{sec:compose}

\begin{figure}[t]
\centering
\includegraphics[width=\linewidth]{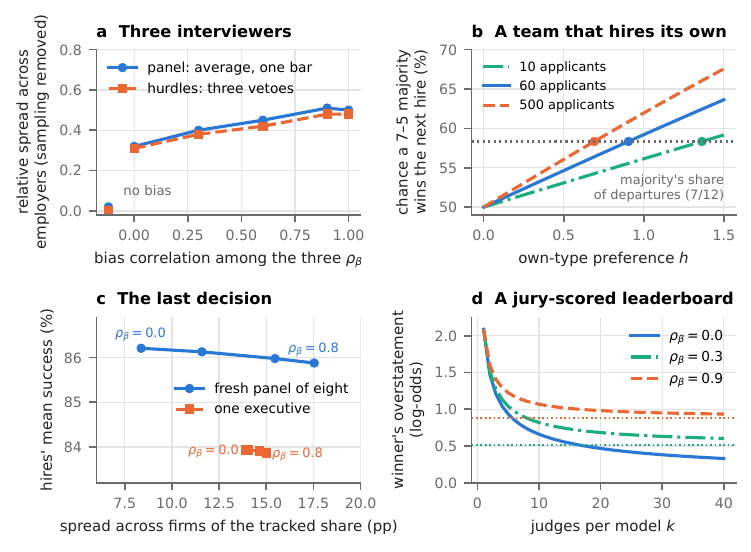}
\caption{\textbf{Four composed systems.} (a) Three interviewers: the relative spread across
employers of the tracked group's share of hires, with sampling removed, rises with the bias
correlation among the three and barely depends on the combining rule. (b) A self-staffing team: the
chance that a $7$--$5$ majority wins the next hire crosses its share of departures ($7/12$) at the
threshold (dots), which falls as the applicant pool grows. (c) The last decision: a fresh panel's
spread across firms, in points of the tracked share, grows with the panels' bias correlation; one
executive's stays near $14$--$15$ points, at about two points less hire quality. (d) A jury-scored
leaderboard of $20$ models: more judges lower the winner's overstatement only to the floor (dotted)
set by what the judges share.}
\label{fig:compose}
\end{figure}

Once gates are wired together, the circuit computes what a design choice does where no single gate
shows it, and in each of the four systems below the answer turns on what the judges share and who
recurs. Three of the systems come from hiring and the fourth from machine learning. Each is computed
at stated parameters (Appendices~\ref{app:perorg}--\ref{app:execfull}), and
Figure~\ref{fig:compose} shows one result from each.

\subsection{Three interviewers: seating sets the spread, the rule buys quality}
\label{sec:2x2}

Three interviewers can combine their judgments in two classic ways~\citep{decorte2006}: as a panel
that averages the three and hires above one bar, or as three hurdles, each interviewer holding a
veto. The more alike the interviewers' biases, the more a candidate's fate depends on which employer
they meet, and the combining rule barely moves that. In the circuit, each employer draws three
interviewers, each with a bias for or against a tracked group whose attribute is unrelated to
quality, and the three biases are correlated at $\rho_\beta$ (Appendix~\ref{app:perorg}). The
measure is how much the tracked group's share of hires varies from employer to employer. Removing
the part that sampling alone would give, and dividing by the group's share of applicants, gives the
\emph{relative spread across employers}. Raising the bias correlation from $0$ to $1$ widens it from
$0.32$ to $0.50$ in the panel and from $0.31$ to $0.48$ in the hurdles (Figure~\ref{fig:compose}a).
The group's share of hires then varies between employers by about a third of its share of
applicants when the interviewers lean independently, and by half when they lean as one.

The combining rule buys hire quality instead, at a price in interviews. The panel's hires have a
mean true success probability $1.2$ to $1.7$ points higher than the hurdles', because the panel
hires on the mean of three noisy judgments and the hurdles in effect on the lowest. The panel pays
with interviews: it interviews every applicant three times, while hurdles that stop at the first
veto interview each applicant about $1.45$ times. This is the compensatory-versus-hurdle trade-off
of personnel selection~\citep{decorte2006}, with the bias now in the interviewers rather than in the
candidates' measured attributes. To make hiring depend less on which employer a candidate meets,
seat interviewers whose biases differ, such as interviewers trained differently, and choose the
combining rule by what an interview costs.

\subsection{A team that hires its own replacements: a threshold of own-type preference}
\label{sec:homophily}

A team that hires its own replacements feeds its makeup back into the panel that hires next. Once members favor their own type strongly enough, a small majority tends to grow at the next
replacement. A larger applicant pool lowers this threshold. Take a team of twelve with two types of member, and let each member add $h$,
the own-type preference, to the judgment of applicants of their own type, so that neither type is
favored overall. At each opening the team averages its members' judgments of sixty applicants and
hires the top one (an \gate{average}, then a \gate{rank} with $s=1$), and the hire replaces a member
who leaves at random.

The threshold comes from a race between two chances. When one type holds seven seats of twelve, it
supplies the departing member with chance $7/12$, or $58.3\%$, and wins the next hire with a chance
that rises with $h$, from $50\%$ at $h=0$ to $59.2\%$ at $h=1$ (Figure~\ref{fig:compose}b). At $7$--$5$ the expected change in the majority's size turns positive at $h=0.905$.
The corresponding threshold from the slope at balance is $h^*=2\sigma_w/m_N=0.898$, where
$\sigma_w$ is the within-type spread of judgments and $m_N$ the expected largest of $N$ standard
normal draws. From $9$--$3$, a team takes a median of $12$ replacements to return to balance at
$h=0$, $56$ at $h=1$ and $1{,}331$ at $h=1.5$ (Appendix~\ref{app:static}).

Three levers move the outcome: the preference itself, the makeup of the hiring panel and the size of
the pool. Lopsided teams grow more common as $h$ rises, below the threshold as well as above it.
After forty replacements, $14.6\%$ of teams are at or past $75/25$ at $h=0$, $38.1\%$ at $h=0.6$ and
$64.8\%$ at $h=1$. So training that lowers the preference helps at every level, and below the $7$--$5$ threshold departures outpace majority hires in expectation at that split. A hiring panel held at balance has no pull toward either type, at any $h$,
because the two types' preferences cancel on it. And the pool moves the threshold itself, from
$1.36$ with ten applicants per opening to $0.90$ with sixty and $0.69$ with five hundred
(Figure~\ref{fig:compose}b). The top of a
larger pool is picked on finer differences in score, so the same preference gives the majority a
larger edge. Chance determines which type first gains seats. Own-type hiring can prolong that lead,
giving the team a reversible form of lock-in (Appendix~\ref{app:catalogue}).

\subsection{The last decision given to one person: a trade-off, not a fix}
\label{sec:exec}

Many pipelines end with one person, a hiring manager, a partner or an editor. Judged as in
\S\ref{sec:2x2}, by how much a tracked group's chances depend on which firm it applies to, handing
that person the last decision is a trade-off, not a fix. After panels whose biases are independent, one person at the last
cut makes the group's chances depend more on the firm. After panels whose biases are strongly
alike, one person makes them depend a little less, but only by making the last cut noisier, so the
firm hires worse. In the circuit, each firm passes its applicants through two panels of eight, whose
members' biases toward the tracked group are correlated at $\rho_\beta$ within a panel, and a fresh
panel of eight, seated the same way, or one executive makes the last cut
(Appendix~\ref{app:execfull}). The measure is the spread across firms of the group's share of hires,
in percentage points.

One person differs from a fresh panel of eight in its bias and in its noise, and the panels' bias
correlation decides which difference counts. By Eq.~\eqref{eq:avg}, a panel's average carries
$\rho_\beta+(1-\rho_\beta)/8$ of one person's bias variance: an eighth when its members lean
independently, but $0.825$ when they are correlated at $0.8$. One person's noise variance is eight
times the panel's at any correlation. A bias spreads firms apart, because it carries the group
across the last bar at some firms and holds it back at others. Noise widens the scores the bar
cuts, so the same bias carries fewer candidates across it, and the cut follows true quality less.
After independent panels the bias that one person adds dominates, and the executive's spread is
$13.9$ points against a fresh panel's $8.4$. After panels correlated at $0.8$, a fresh panel already
carries nearly one person's bias, so one person adds mostly noise. The executive's spread is
narrower, $15.0$ points against $17.5$, and its hires succeed $83.9\%$ of the time against $85.9\%$
(Figure~\ref{fig:compose}c). Crossing the two on the same survivors confirms the split. One person's
bias alone widens the spread and costs under half a point of success, and one person's noise alone
narrows it and costs about two points (Appendix~\ref{app:execfull}). Decorrelating the panels narrows the
spread at no cost: a fresh panel after independent panels gives the $8.4$ points above, and the
best hires, $86.2\%$.

\subsection{A leaderboard scored by a jury: a floor more judges cannot lower}
\label{sec:jury}

A leaderboard that scores models with a jury of model judges can lower the overstatement of its
winner by adding judges only down to a floor set by what the judges share, and a jury that shares
less lowers the floor where more of the same judges cannot. Let $N$ equally good models each be
scored by the mean of $k$ judges, an \gate{average} per model, and let the top score win, a
\gate{max} over models. Each judge's bias toward a given model, a taste for its style or its family,
differs from model to model, with scale $\sigma_\beta$ and correlation $\rho_\beta$ across judges, so
$\sigma_\beta^2\rho_\beta$ is everything the judges hold alike about that model, the common part and
the panel part together. The \gate{max} selects on the whole averaged error of Eq.~\eqref{eq:avg}, so
the winner is overstated by
\begin{equation}
m_N\sqrt{\sigma_\beta^2\!\left[\rho_\beta+\frac{1-\rho_\beta}{k}\right]+\frac{\sigma_\varepsilon^2}{k}},
\label{eq:jury}
\end{equation}
which falls with $k$ only to the floor $m_N\sigma_\beta\sqrt{\rho_\beta}$, set by what the judges
share. With $20$ models, $\sigma_\beta=0.5$ and $\sigma_\varepsilon=1$, one judge overstates the
winner by $2.09$ log-odds. Ten judges cut that to $0.66$ if their tastes are uncorrelated, but only
to $1.07$ at $\rho_\beta=0.9$, against a floor of $0.89$ that no number of such judges goes below
(Figure~\ref{fig:compose}d).

Because the floor depends on $\sigma_\beta^2\rho_\beta$, choosing a jury trades the size of the
judges' biases against how alike they are. Judges from another model family lower the floor only if
they are less alike without being much more biased. When the models also differ in true quality,
the overstatement shrinks by the factor $\sqrt{1-\lambda}$, with $\lambda$ computed from the
averaged error, so part of the winning margin is real.

\section{What it means for machine learning}
\label{sec:ml}

For machine learning the circuit says three things. Its evaluation pipelines inherit the failures of
the gates they are built from. Because the catalogue's phenomena need so little of the evaluator, an
agent simulation of people that reproduces them shows little. And model judges allow what human
panels rarely do, a measurement of what the judges share.

\gate{max} is model selection, best-of-$n$
sampling and the leaderboard's winner; \gate{average} is the ensemble and the jury of judges;
\gate{threshold} is a classification cutoff; \gate{rank} is fixed-size retrieval. The best of many
checkpoints on one validation set is the winner of \S\ref{sec:max}, and a leaderboard scored by
model judges is the system of \S\ref{sec:jury}, so such pipelines can be redesigned with the circuit
before they are run.

Research that simulates people with language-model agents~\citep{park2023generative} meets the
catalogue of \S\ref{sec:reach} from the other side. Every phenomenon in Table~\ref{tab:catalogue}
arises in the circuit from an evaluator with nothing but a bias and noise, so a population of agents
that reproduces one shows little about whether its agents judge like people. Reproducing the wisdom of
crowds shows only that a simulated population has independent noise, and larger aligned models lack
even that, giving the exact answer with no spread~\citep{aher2023}. Matching a human crowd's
accuracy shows no more. Twelve models matched $925$ human forecasters while sharing an acquiescence bias that averaging does not remove~\citep{schoenegger2024silicon}.
A sharper test is a change of arrangement on which the circuit and the agents disagree. Make the
change, and see whether people follow the agents or the circuit.

Every design choice in \S\ref{sec:compose} turns on what the judges share, and agreement between
panels cannot isolate it. Two evaluators who judge the same case differ by their own parts and their
noise, while what they share cancels, so a noise audit~\citep{kahneman2021} cannot see it; two
independently staffed panels agree more when they share a bias, so a consistency experiment counts
what both share as agreement, the old point that reliability is not
validity~\citep{brennan2001generalizability} (Appendix~\ref{app:deriv}). For ensembles of model
judges, too, the shared error is not identifiable from their scores~\citep{afrin2026evaluator}.
Repeated scores on cases with known answers do isolate it. Repeating a judge's score on one case
separates its noise from its bias, the known answer separates its bias from the case's quality, and
comparing judges case by case gives what they share, as models' error correlations are already
measured on labeled benchmarks~\citep{kim2025correlated}. Then the model can be tested directly:
predict what replacing a judge does to a leaderboard's overstatement, then replace it.

\section{Related work}
\label{sec:related}

Engineering decisions has named precedents in decision intelligence~\citep{pratt2019link} and
market design~\citep{roth2002economist}; DSE engineers a different object. Four lines of work come
closest, and the
circuit meets several of them as special cases or at one composition. None builds the errors from who sits where and carries them through four gate rules, composed
stages and feedback staffing (Appendix~\ref{app:related}).

\paragraph{Designing arrangements of fallible agents.} \citet{sah1986} wire fallible screens in
series and in parallel and compare their errors, and \citet{christensen2010} treat the full range of
such architectures, with the most reliable designs for a given number of agents. Their screens'
errors are independent across screens and across candidates. With no bias and a probit screen the
circuit returns Sah and Stiglitz's error rates.

\paragraph{Pricing a design at one stage.} Generalizability theory's decision studies compute how a
rating's dependability changes with the number of raters and occasions~\citep{brennan2001generalizability}.
A decision study of multiple mini-interviews found that more interviews give more consistent results
than more interviewers per interview~\citep{eva2004relationship}. Multistage selection computes a
design's quality and adverse impact across stages~\citep{decorte2006}, and \S\ref{sec:2x2} reproduces
its compensatory-versus-hurdle trade-off with the bias in the interviewers.

\paragraph{Correlated judges at one gate.} Correlated experts add little to an
average~\citep{hogarth1978}. In large forecasting groups, the type of forecaster with the least
within-type covariance should dominate~\citep{lamberson2012optimal}, the same rule as the
leaderboard's floor. Language models' errors are measured to be correlated~\citep{kim2025correlated},
and for ensembles of model judges the mean keeps the error the judges share, with bounds on
best-of-$K$ overstatement~\citep{afrin2026evaluator}, the leaderboard's floor at one composition.

\paragraph{Groups that admit their own.} In a growing group whose members admit newcomers by
consensus, there is a critical noise level below which the group stays
cohesive~\citep{fenoaltea2023phase}. Affinity bias under changing committees has been posed as a
bandit problem with only biased feedback observed~\citep{faw2025affinity}, and hiring strategies have
been compared in a committee with similarity biases~\citep{morgan2014comparing}. The team's threshold
in \S\ref{sec:homophily} is of this kind; the circuit derives it from its gates and shows how the
applicant pool moves it.

\section{Conclusion and limits}
\label{sec:conclusion}

Decision System Engineering treats a decision system as something to write down and compute before
it is built, and the decision circuit is its first model. It asks little of the evaluator, only a
bias that stays with them and noise drawn afresh, yet with four kinds of gate and any wiring it
gives mechanisms for $31$ reported patterns and computes the effects of specific redesigns. At the parameters computed
here, its answers are concrete. Seat interviewers whose biases differ to make hiring depend less on
the employer. Keep a self-staffing team's own-type preference below a threshold that falls as the
applicant pool grows, or hold its hiring panel at balance. Treat one person at the last cut as a
trade-off, not a fix: after alike panels that person makes hiring depend less on the firm only by
hiring worse, while decorrelating the panels does it at no cost. And lower a leaderboard's floor by
choosing judges who share less, not by adding judges. For machine learning, a pipeline can be
redesigned before it runs, and an agent simulation that reproduces these phenomena has shown only
what a bias and noise show. Anyone who runs a pipeline, a panel or an
evaluation stack can ask at every gate who recurs there and what the judges share, and then compute,
for stated parameters, what a change will do.

\paragraph{Limits.} The model's accuracy waits on three things. No bias parameter has been estimated
from a real panel; \S\ref{sec:ml} says how repeated scores from model judges would supply them. The
evaluators do not communicate, strategize, update or learn, and the gates do not include ratings by
pairwise comparison, such as arena leaderboards; phenomena that need these lie outside this first
model. And one system, peer review, has been compared with measurements; prospective tests of
predicted design changes would show how accurate the circuit is.

\bibliographystyle{plainnat}
\bibliography{references}

\appendix

\section{Extended related work}
\label{app:related}

This appendix sets the decision circuit against the closest work in each line. For each line it says
what the circuit inherits and where it differs; a fuller annotated bibliography
of the sources behind each catalogue row will be released with the code.

\paragraph{A named discipline with precedents.} The string ``decision system engineering'' is not in
use as a defined field, but the idea of engineering decisions is not new, and the paper does not
claim it. Decision intelligence engineers an organization's causal model of how actions lead to
outcomes~\citep{pratt2019link}. Market design treats the design of markets as engineering, with
strategic agents~\citep{roth2002economist}. DSE engineers a different object: the arrangement of
fallible evaluators who judge candidates (who sits at which stage, under which rule, with which bias
correlation), designed from an explicit model of their errors whose outcome is computed before the
system is built. Designing such arrangements is an old aim. \citet[p.~716]{sah1986} describe how
``the constituent decision-making units are arranged together in a system'', and draw their series
and parallel screens ``from the design of relay circuits''~\citep[p.~718]{sah1986}; ``decision
circuit'' continues that analogy.

\paragraph{Designing arrangements of fallible agents.} \citet{sah1986} compare hierarchies, which
reject more good projects, with polyarchies, which accept more bad ones, and present them as
``building blocks for complex organizations''. Their model carries one error term, independent across
screens and, in their own statement, across projects. Their screens may take any screening function;
with $\beta=0$ and a probit screen the circuit returns their equation exactly
(Appendix~\ref{app:deriv}), so the circuit generalizes their model along the bias and specializes it
along the screening function. \citet{christensen2010}, starting from the premise that individuals in
an organization are fallible, treat the full range of architectures between hierarchy and polyarchy
and derive the most reliable structures for a given number of agents; their screens too are
independent and binary. A shared bias couples the screens' decisions on one candidate, and a
persistent bias couples one screen's decisions on similar candidates. The circuit adds the
persistent bias and its correlation, the \gate{average}, \gate{max} and \gate{rank} rules on
continuous scores, populations that pass from gate to gate, and gates staffed by their own output.

\paragraph{Pricing a design at one stage.} Generalizability theory splits a rating's variance into
facets, among them the rater and the occasion, and its decision studies compute, before a design is
used, how dependability changes with the number of raters and
occasions~\citep{brennan2001generalizability}. That is the \gate{average} gate's measurement-design
ancestor: Eq.~\eqref{eq:avg} with a rater facet and an occasion facet. In the same framework, rater
variance shared across a design enters the universe score, the generalizability analogue of
true-score variance, which is why agreement cannot reveal the part of a bias that the compared raters
share (\S\ref{sec:ml}); the circuit inherits that point. \citet{eva2004relationship} ran a nine-station
multiple mini-interview with $54$ candidates, in which three stations had two faculty interviewers,
three had two community members and three had one of each. Overall reliability was $0.78$; a
decision study concluded that more stations give more consistent results than more interviewers per
station; and ratings diverged most for faculty--community pairs, a measured effect of who sits with
whom at one gate. The parts of the bias have also been measured in the field.
\citet{scullen2000understanding} model performance ratings as ratee effects, idiosyncratic rater
effects, rater-perspective effects (boss, peer, subordinate) and random error; in two samples of
managers rated by seven raters each, idiosyncratic rater effects accounted for over half of the
rating variance ($62\%$ and $53\%$), and the perspective effects, which raters of one kind share,
were small. That is a measurement of the own part of the bias and of a part held alike by raters
of one kind.

\paragraph{Stages composed in personnel selection.} \citet{decorte2006} compute analytically the mean
performance and adverse impact of multistage selection from mixtures of populations, with threshold
stages and compensatory composites. The three-interviewer circuit of \S\ref{sec:2x2} is this
literature's comparison of compensatory and multiple-hurdle designs. The two differ in where the group
difference lives: there, in predictors, fixed; here, in evaluators, drawn per employer, which
separates the spread across employers from the pooled share of all employers together. Part of
Appendix~\ref{app:perorg} is therefore not new. A bias common to every evaluator is, seen from the
predictor, test bias in the sense of \citet{cleary1968test}, who defined a test as biased for a
subgroup when it makes consistent nonzero errors in predicting that subgroup's outcome; fed through
multistage models, such a predictor gives the common part's effect on pooled representation. With
fixed evaluators and Gaussian errors, a series of \gate{threshold} gates is the multivariate-normal
selection these studies compute. They do not give the spread across employers and its dependence on
the bias correlation among the evaluators an employer happens to draw, nor \gate{max}, \gate{rank}
or gates staffed by their own output. The closest recent work on composition treats an agentic
discovery pipeline as a sequence of filtering stages and analyses it with classical multistage
selection theory under a truncated multivariate normal: error shared across components costs yield,
and which stage to improve depends on the funnel's geometry~\citep{puttagunta2026component}. Its
shared error belongs to the pipeline's components; in the circuit it belongs to evaluators and
follows from who sits together.

\paragraph{Correlated judges at one gate.} That independent errors wash out under averaging goes back
to \citet{galton1907}, and the aggregation of correlated judgments is old too: \citet{hogarth1978}
gives the validity of a mean of correlated experts. The exact variance of an average of equicorrelated
terms is standard~\citep[\S15.2]{hastie2009elements}. \citet{lamberson2012optimal} characterize the
optimal composition of a forecasting group in which forecasters of one type have higher covariance
than forecasters of different types: in small groups the most accurate type should be in the
majority, while in large groups the type with the least within-type covariance should dominate. The
leaderboard's rule in \S\ref{sec:jury}, that past a few judges what the judges share,
$\sigma_\beta^2\rho_\beta$, decides, is the same rule at a \gate{max} downstream of the average.
\citet{kahneman2021} separate bias from level, pattern and occasion noise, and the circuit maps onto
their terms exactly (\S\ref{sec:circuit}). For language models, \citet{kim2025correlated} measure
correlated errors across many models, and \citet{afrin2026evaluator} show for ensembles of model
judges that the mean keeps the common-mode error, that this error is not identifiable from the
judges' scores, and that best-of-$K$ overstatement can be bounded. Each of these concerns one gate.
The circuit's addition is to build the errors from who sits where and carry them through gates of
four kinds and through feedback.

\paragraph{Groups that admit their own.} \citet{kanter1977} described homosocial reproduction in a
firm, and models of increasing returns~\citep{arthur1989} give the reinforcement its mathematics
without placing it in who evaluates. \citet{fenoaltea2023phase} study a group that grows by a noisy
admission process with two types of member, cohesion being the fraction of members of one type. When
one member decides each admission, growth lowers cohesion even at very low noise; when admission
needs a consensus, there is a critical noise level below which the group stays cohesive. In hiring
and admissions, today's selected candidates later sit on the committees that decide;
\citet{faw2025affinity} model a bias that depends on how common a candidate's traits are among the
committee, and pose the problem as a bandit with only biased feedback observed.
\citet{morgan2014comparing} compare hiring strategies in a committee with similarity biases. The
team's threshold of \S\ref{sec:homophily} is of the same kind as Fenoaltea et al.'s critical noise
level. The circuit derives it from an \gate{average} and a \gate{rank} gate, computes it forward from
stated parameters, and shows how it moves with the size of the applicant pool.

\paragraph{One gate's law in each literature.} The \gate{max} law is the optimizer's curse and
postdecision disappointment~\citep{smith2006optimizer} and the auction winner's
curse~\citep{thaler1988}; its precision form appears in rankings of small
units~\citep{wainer2007,gelman1999}. The \gate{threshold} law is Becker's outcome
test~\citep{becker1957}, with its infra-marginality problem~\citep{simoiu2017}. It is also Berkson's
collider~\citep{berkson1946} and statistical discrimination~\citep{aigner1977}, whose Bayesian
evaluator reverses the circuit's sign. The circuit claims none of these laws; it claims that one
evaluator model produces all of them and lets them compose.

\paragraph{Fairness and monoculture.} \citet{emelianov2020} study noise that differs by group at one
gate. \citet{kleinberg2021monoculture} show that a shared algorithmic ranking across firms can lower
welfare, and \citet{kim2025correlated} measure the systemic exclusion that follows when firms screen
with language models. The circuit computes outcomes forward from error terms, separates the bias from
the noise in the correlation across firms, and composes.

\paragraph{Peer review as a measured system.} The NeurIPS consistency
experiments~\citep{cortes2021,beygelzimer2023} and an NSF re-review~\citep{cole1981} measured how often
independent committees disagree. Persistent per-reviewer bias is already modeled there:
\citet{cortes2021} calibrate the 2014 scores with a Gaussian-process variant of the Platt--Burges
model, which writes a reviewer's score as $r_{ij}=g_i+b_j+e_{ij}$, with an additive bias $b_j$ for
reviewer $j$. These models estimate each reviewer's bias at one aggregation gate. The circuit's
treatment (\S\ref{sec:reach}) is a consistency check of the \gate{average}~$\to$~\gate{threshold}
structure with one measured input, in which no bias term enters, and one prediction from outside
reliability data.

\paragraph{Machine-learning evaluation as a circuit.} Model selection overfits through the variance of
its criterion~\citep{cawley2010overfitting}. Best-of-$n$ optimization against a proxy reward selects
for noise, the regressional form of Goodhart's law, and its gold reward later declines, which a
Gaussian \gate{max} cannot produce~\citep{gao2023scaling}. Language-model judges have position and
verbosity biases~\citep{zheng2023judging}, and juries of diverse judges reduce such
biases~\citep{verga2024juries}; models' errors are correlated, and a judge inflates the accuracy of
models less accurate than itself~\citep{kim2025correlated}. The jury-scored leaderboard of
\S\ref{sec:jury} composes these: an \gate{average} over judges whose biases are correlated, followed
by a \gate{max} over models. Classifier cascades pass survivors from stage to
stage~\citep{viola2001rapid}, composed selection stages without an evaluator error model.

\paragraph{Simulating people with language models.} Generative agents~\citep{park2023generative} and
other simulated populations are checked against human results. \citet{aher2023} report a
hyper-accuracy distortion in larger aligned models, and \citet{schoenegger2024silicon} find an
ensemble of models as accurate as a human crowd but with an acquiescence bias. The circuit offers this
literature a simple point of comparison: a phenomenon that an evaluator with only a bias and noise
reproduces at the right gate is weak evidence of human fidelity, though an agent population could
match finer features that such an evaluator misses.

\section{The catalogue in full}
\label{app:catalogue}

Each entry gives the documentation (numbers as reported), the circuit motif, and the support used in
Table~\ref{tab:catalogue}: \textbf{Mag}, magnitudes compared with measured values (peer review
only); \textbf{Dir}, the documented direction shown at stated parameters, by simulation or, in one
row, by a closed-form illustration; \textbf{Cal}, one parameter calibrated to one number, which is
not a test; \textbf{CF}, derived in closed form, including re-derivations of known results;
\textbf{Map}, a sufficient mechanism for the pattern with no fit. The table holds $31$ documented
phenomena: $1$ Mag, $9$ Dir, $3$ Cal, $8$ CF and $10$ Map. The mark $\dagger$ identifies a persistent or shared
evaluator preference in the proposed explanation. Several such mechanisms already appear in the
respective literatures, including taste-based bias~\citep{becker1957} and the position bias of
language-model judges~\citep{zheng2023judging}. Appendix~\ref{app:deriv} gives the gate calculations.

\paragraph{\gate{average}.}
\emph{Wisdom of crowds} (Dir): independent errors average away~\citep{galton1907}; across people, the
error of an average falls along a hyperbola toward ``the bias of the distribution of beliefs in the
population''~\citep{vul2008}. Motif: Eq.~\eqref{eq:avg} at $\rho\approx0$, with the floor as the
asymptote; Figure~\ref{fig:avg}.
\emph{Committee paradox} (Dir, $\dagger$): the validity of a mean of experts is limited by their
intercorrelation~\citep{hogarth1978}. The source is theoretical; the row re-expresses its result as
the floor at $\rho_\beta$ and $k_{\mathrm{eff}}$.
\emph{The crowd within} (Cal, $\dagger$): among $428$ participants, two guesses from one person were
worth $1.11$ people at once and $1.32$ after three weeks~\citep{vul2008}, numbers of different people
whose averaged guesses would match the error of the two own guesses, read off the fitted hyperbola.
Motif: an \gate{average} of two occasions of one evaluator, $\rho_\beta=1$; occasion noise's share of
a person's own error is at least $0.20$ for immediate guesses and at least $0.49$ after three weeks
(Appendix~\ref{app:deriv}); whatever is not occasion noise is a personal bias that only other people
average away. This calibrates one parameter to one number.
\emph{Bootstrapping} (Map): a linear model of a judge's own judgments can predict better than the
judge, and a meta-analysis of lens-model studies analyzes when~\citep{karelaia2008}. Motif: removing
$\varepsilon$, assuming that the model of the judge reproduces the judge's view, $\lgt(q)+\beta^\top
x$, exactly. A lens-model model of the judge regresses judgments on cues~\citep{karelaia2008}, while
Eq.~\eqref{eq:model} lets the judge see true quality itself, so the closed form
(Appendix~\ref{app:deriv}) shows a sufficient mechanism under that assumption, and its prediction
that bootstrapping always helps follows from the setup rather than from the data. The cases where it
does not help are out of reach.
\emph{Large differences between judges} (Cal): in an insurance audit, the median difference between
two underwriters' premiums was $55\%$ and between two claims adjusters' assessments $43\%$, against
$10\%$ expected by executives~\citep{kahneman2021}. Motif: two single evaluators, whose difference
carries the varying bias $\delta$ and the noise; calibrated in Appendix~\ref{app:deriv},
illustratively, because premiums are not log-odds. The circuit gives the size of the differences; the
executives' underestimate is a fact about beliefs that it does not produce.
\emph{LLM juries} (Map, $\dagger$): on Arena-Hard, a three-family panel that averaged the
judges' grades produced a ranking with Pearson correlation $0.917$ and Kendall's $\tau=0.778$
against the human leaderboard. For the GPT-4 judge the corresponding values were $0.817$ and
$0.667$~\citep[Table~2]{verga2024juries}. In the circuit, a preference for one's own model
family enters the mean with weight $1/k$ when only one of $k$ judges shares it. A preference
shared by all judges survives averaging. The observed ranking comparison illustrates the pattern.
\emph{Swap-averaging} (CF, $\dagger$): position consistency under swapped order was $65.0\%$ for
GPT-4, $46.2\%$ for GPT-3.5 and $23.8\%$ for Claude-v1, and the remedy calls the judge twice with the
positions swapped~\citep{zheng2023judging}. Motif: a fixed bias on position whose sign the swap
flips, averaged at $k=2$ with perfectly anti-correlated bias; verbosity has no sign flip and stays at
the $\rho_\beta=1$ floor. The remedy is that of \citet{zheng2023judging}, and the derivation is
arithmetic; the row re-expresses it as a case of Eq.~\eqref{eq:avg}.
\emph{Silicon crowd} (Map, $\dagger$): an ensemble of $12$ models forecasting $31$ binary questions
was ``statistically indistinguishable'' from $925$ human forecasters, with mean predictions
significantly above $50\%$~\citep{schoenegger2024silicon}. Motif: noise averaged away, a shared
acquiescence bias at the floor.
\emph{Hyper-accuracy} (Map): larger aligned models give ``inhumanly accurate answers'' with no
spread~\citep{aher2023}. Motif: an \gate{average} with no independent noise, which aggregation cannot
improve.

\paragraph{\gate{max}.}
\emph{Winner's curse} (Dir): a review of the curse reports that in the 1969 Alaska North Slope lease
sale the winning bids summed to \$900M against \$370M for the second-highest; that in classroom
auctions of a jar worth \$8.00 the mean winning bid was \$10.01; that groups of six or seven bidders
lost money where groups of three or four earned part of the predicted profit; and that takeover
bidders may overpay out of hubris~\citep{thaler1988}. Motif: Eq.~\eqref{eq:identity},
$\sigma_\varepsilon m_N$; the circuit produces the direction and its growth with the number of
bidders, and these documented magnitudes are not compared with its output. With different biases the
winner is also selected on bias, which is the hubris channel~\citep{thaler1988}. The curse arises
because the evaluator does not condition on winning; whether bidders shade their bids enough is a
strategic question, out of reach.
\emph{Optimizer's curse} (Dir): choosing the best of noisy estimates yields expected
disappointment~\citep{smith2006optimizer}. Motif: as above.
\emph{Model selection} (CF): the variance of a selection criterion causes over-fitting in model
selection that can rival the differences between algorithms~\citep{cawley2010overfitting}. Motif:
\gate{max} over items, $\sqrt{V(1-\lambda)}\,m_N$, with $V$ the variance of the score's whole error.
\emph{Best-of-$n$ overoptimization} (Map): proxy reward keeps rising while gold reward peaks and
falls~\citep{gao2023scaling}. Motif: the Gaussian \gate{max} reproduces the regressional term (gold
linear in $\sqrt{\log n}$, with slope proportional to $\lambda$), not the fall, which a Gaussian
\gate{max} cannot produce.
\emph{Small units at both ends} (Dir): of Pennsylvania's top-scoring $50$ schools, six were among the
smallest $3\%$, an over-representation by a factor of four~\citep{wainer2007}; very high and very low
rates appear disproportionately in poorly sampled areas~\citep{gelman1999}. Motif: unequal
$\sigma_\varepsilon$ at a \gate{max} or \gate{rank} (\S\ref{sec:max}). An evaluator that shrinks noisy
estimates toward the mean reverses the precision result~\citep{gelman1999}; the circuit's evaluator
does not shrink.

\paragraph{\gate{threshold}.}
\emph{Outcome test} (Dir, $\dagger$): bias is read from the outcomes of those
admitted~\citep{becker1957}; within investor, the female-led startups male investors show interest in
outperform the male-led ones they show interest in~\citep{ewens2020}. Motif: Eq.~\eqref{eq:gap}; the
bias $b$ is held by the evaluator.
\emph{Infra-marginality} (CF): average outcomes need not reflect the margin, and in North Carolina
traffic stops outcome and threshold tests disagree in direction~\citep{simoiu2017}. Motif: a closed
form for groups with different quality distributions (Appendix~\ref{app:deriv}), in which admits'
averages differ with no bias at all. The paper's simulation has one shared quality distribution, so
its averages ($0.341$ against $0.177$) and its gap $2b\lambda$ agree in sign and do not show the
reversal.
\emph{Berkson's paradox} (CF): selection into a sample induces spurious
associations~\citep{berkson1946}. Motif: a negative covariance of true quality and noise among the
admitted.
\emph{Group-dependent noise} (CF): with noise that differs by group, a group-oblivious selector
under-represents the low-variance group~\citep{emelianov2020}. In contrast, with a Bayesian employer,
differential reliability alone does not lower a group's average wage~\citep{aigner1977}. Motif: a
raw-score bar over-admits the noisier group and admits worse members of it; a Bayesian evaluator
reverses the sign.
\emph{Regression after selection} (CF): decline after promotion follows from regression to the
mean~\citep{lazear2004}, and postdecision surprises lean toward
disappointment~\citep{smith2006optimizer}. Motif: the shortfall $(1-\lambda)(y-\mu)+\lambda c$, a
statistical identity the circuit re-derives.
\emph{Hierarchy and polyarchy} (CF): hierarchies reject more good projects and polyarchies accept
more bad ones~\citep{sah1986}. Motif: the special case without bias, with a probit screen
(Appendix~\ref{app:deriv}). Recovering the model of \citet{sah1986} checks the generalization rather
than extending its reach.

\paragraph{Peer review.}
\emph{Committees disagree} (Mag): the 2014 NIPS committees decided $43$ of $166$ duplicated papers
differently, $25.9\%$, and $49.5\%$ of one committee's accepts were rejected by the
other~\citep{cortes2021}; the two NeurIPS 2021 committees disagreed on $23.0\%$ of $882$ papers,
$50.6\%$ of accepts were rejected by the other committee, and ``making the conference more selective
would increase the arbitrariness''~\citep{beygelzimer2023}; when $150$ NSF proposals in three programs
were re-reviewed by new reviewers, getting a grant was found to depend ``to a significant extent on
chance''~\citep{cole1981}. Motif: \S\ref{sec:reach}. The committees' score correlation is an input,
each year is cut at its own acceptance rate, and disagreement and the outside share are one
statistic, so this row is one consistency check of the \gate{average} $\to$ \gate{threshold}
structure. No bias term enters it, so it says nothing about the persistent bias. The selectivity
claim is reproduced in closed form and holds under any bivariate-normal model with a hard cut; the
median-cut comparison with the NSF re-review is approximate (Appendix~\ref{app:deriv}).
\emph{Agreement from reliability} (Cal): a meta-analysis of $70$ reliability coefficients from $48$
studies puts mean inter-rater reliability at $0.34$~\citep{bornmann2010}. That value pools several
kinds of reliability measure, so the row uses it only as an illustrative single-reviewer value.
Motif: Eq.~\eqref{eq:avg} with independent noise, applied to reliability, at three or four reviewers
per panel. The predictions, $0.61$ and $0.67$, exceed the correlations measured at NeurIPS ($0.55$ to
$0.586$); the first lies within the NSF re-review's $0.60$--$0.66$~\citep[p.~882]{cole1981}. This
calibrates one parameter to one number.
\emph{Reviewer agreement after selection} (Dir): in a mock NIH review, $43$ primary reviewers
gave $83$ preliminary ratings to $25$ application versions. Sixteen had been funded on the first
submission and nine were the original versions of applications funded after revision. The
estimated intraclass correlation (ICC) was $0$ (95\% CI $[0,0.14]$)~\citep{pier2018}. All ratings
came from this selected pool. In a separate sample of $2{,}076$ NIH applications, the single-rater
ICC was $0.34$ overall and its restricted maximum-likelihood estimate reached zero in top-ranked
subsets~\citep{erosheva2021}. Those subsets were ranked by the same reviewers' mean score.
Equation~\eqref{eq:fresh-review} gives the direction when separate selectors and reviewers
observe the same underlying quality with independent errors. Neither study measures that exact
before-and-after comparison.

\paragraph{Composition and feedback.}
\emph{Compensatory and multiple hurdle} (Dir): \S\ref{sec:2x2} and Appendix~\ref{app:perorg}; the
selection studies are described in Appendix~\ref{app:related}. Adverse impact there comes from
predictor differences, so the row carries no $\dagger$.
\emph{Unstructured interviews} (Map, $\dagger$): \citet{schmidt1998} reported predictive
validities of $0.51$ for structured interviews and $0.38$ for unstructured interviews. At fixed
panel size, a common scoring rubric can remove cue-weight deviations that survive averaging and
lower correlation with the performance criterion (Eq.~\eqref{eq:cue-validity}).
\emph{Powerful chief executives} (Map, $\dagger$): \citet{adams2005} found greater variation in
firm performance where CEOs held greater decision authority. In the circuit a single decision
maker can shift a firm's selections toward projects that person favors. With equal stakes,
selected projects' mean success probability determines expected portfolio payoff.
Appendix~\ref{app:execfull} computes the spread of that payoff across firms. The comparison
addresses the cross-sectional pattern in the study.
\emph{Leaky pipeline, evaluator part} (Dir, $\dagger$): per-transition gender differences in academic
careers are dwindling but nonzero~\citep{shaw2012}. Motif: a fixed series of biased gates
(Appendix~\ref{app:static}). The candidates' own choices, the main bottlenecks there, are out of
reach, so the row covers the evaluators' part only.
\emph{Cascades} (Map): examples that pass the first stages of a detection cascade are
harder~\citep{viola2001rapid}. Motif: survival compresses true quality and lowers $\lambda$
downstream.
\emph{Monoculture} (CF, $\dagger$): a shared ranking across firms can lower
welfare~\citep{kleinberg2021monoculture}; with firms each screening r\'esum\'es with a language model,
about $20\%$ of applicants were systemically excluded at $20$ firms~\citep{kim2025correlated}. Motif:
parallel \gate{threshold} gates correlated across firms; the strategic equilibrium is out of reach.
\emph{Homosocial reproduction} (Map, $\dagger$): \citet{kanter1977}; \S\ref{sec:homophily}. The
documentation is qualitative, so the circuit shows sufficiency, not fit.
\emph{Long-lived majorities} (Map, $\dagger$): \citet{arthur1989} modeled path dependence
under increasing returns. In an experimental music market, an imposed popularity ranking shifted
subsequent choices and some songs later recovered~\citep{salganik2008}. The team circuit supplies
a staffing analogue. An early lead gives its type greater influence over the next hire. At any
finite preference a reversal remains possible. Appendix~\ref{app:static} computes how long a lead
lasts under the team's assumptions.

\paragraph{Results that are not documented phenomena.} Some results the circuit produces are not in
the table because no source documents them. That a noise audit cannot see what the judges share,
and that a consistency experiment counts what both panels share as agreement, is derived
(\S\ref{sec:ml}, Appendix~\ref{app:deriv}); it restates for decision systems the point that
reliability is not validity~\citep{brennan2001generalizability}. That a batch selected by \gate{rank}
keeps its count as the pool drifts (\S\ref{sec:rank}) is a design regularity for which no empirical
source was found. The four compositions of \S\ref{sec:compose} are computed results.

\paragraph{Out of reach.} Phenomena that need evaluators to infer from one another or to communicate,
to respond to incentives, to judge their own judgment, or to learn or re-estimate from earlier rounds
are out of reach; communication that makes judgments alike enters the circuit only as a higher
correlation. So is an aggregate gap that disappears gate by gate when candidates choose among parallel
gates of different selectivity, since that choice is not one of the four rules; and so is feedback
from a decision to the outcome it is scored against, since the circuit scores decisions against a
fixed true quality.

\section{Derivations}
\label{app:deriv}

\begin{figure}[h]
\centering
\includegraphics[width=0.86\linewidth]{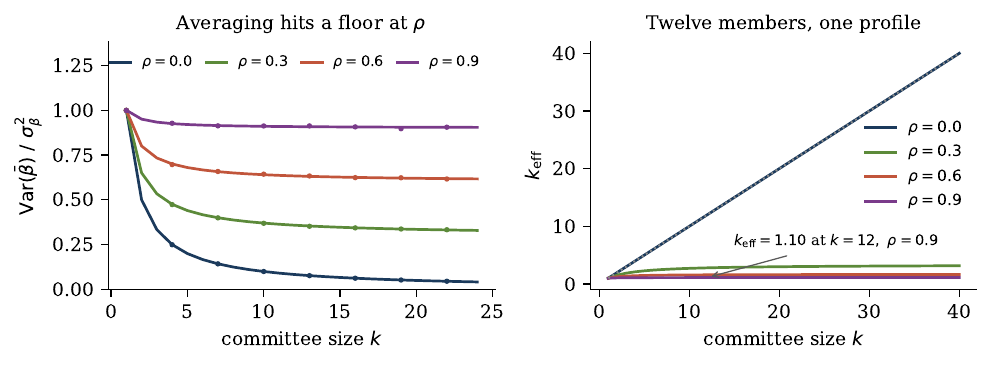}
\caption{\textbf{The \gate{average} gate.} Left: variance of the averaged bias against committee
size; lines are Eq.~\eqref{eq:avg}, markers simulation; $\rho$ in the legend is $\rho_\beta$. The
curve flattens at $\rho_\beta$. Right: effective committee size. At $\rho_\beta=0.9$, twelve members
are worth $1.10$.}
\label{fig:avg}
\end{figure}

\paragraph{The variance of an average and $k_{\mathrm{eff}}$.} For $k$ terms with common variance
$\sigma^2$ and pairwise correlation $\rho$, $\Var(\bar v)=k^{-2}[k\sigma^2+k(k-1)\rho\sigma^2]
=\sigma^2[\rho+(1-\rho)/k]$, which is Eq.~\eqref{eq:avg}. Setting $\sigma^2/k_{\mathrm{eff}}$ equal to
it gives $k_{\mathrm{eff}}=k/[1+(k-1)\rho]$. At $k=12$, $\rho=0.9$: $12/10.9=1.10$. The averaged noise
of twelve independent judgments has standard deviation $1/\sqrt{12}=0.29$ of one judgment's. A common
bias $\bar\beta$ adds the same constant to every term and is not averaged at all
(Figure~\ref{fig:avg}).

\paragraph{The special case without bias.} With $\beta=0$, independent Gaussian noise and a probit
screen, a screen passes quality $L$ with $p=\Phi((L-\tau)/\sigma_\varepsilon)$. Two screens in series
pass with $p^2$; ``either of two'' passes with $1-(1-p)^2=p(2-p)$. These are Eq.~(2) of
\citet{sah1986}, whose screens allow any screening function $p(L)$; the circuit fixes it to the
probit. Their equation needs decisions independent across screens and across candidates.

\paragraph{$j$-of-$n$ votes.} ``At least $j$ of $n$ judgments clear $\tau$'' holds exactly when the
$j$-th largest judgment clears $\tau$, so ``either of two screens suffices'' is $j=1$, unanimity is
$j=n$ and majority is the median. With independent noise and no bias,
$\Pr(\text{accept}\mid L)=\sum_{i\ge j}\binom{n}{i}p^i(1-p)^{n-i}$, with a probit $p$. With
$\rho_\beta>0$ the judgments are exchangeable but not independent and the binomial formula fails;
the circuit then computes the vote by simulation. No vote is computed in this paper.

\begin{figure}[h]
\centering
\includegraphics[width=0.86\linewidth]{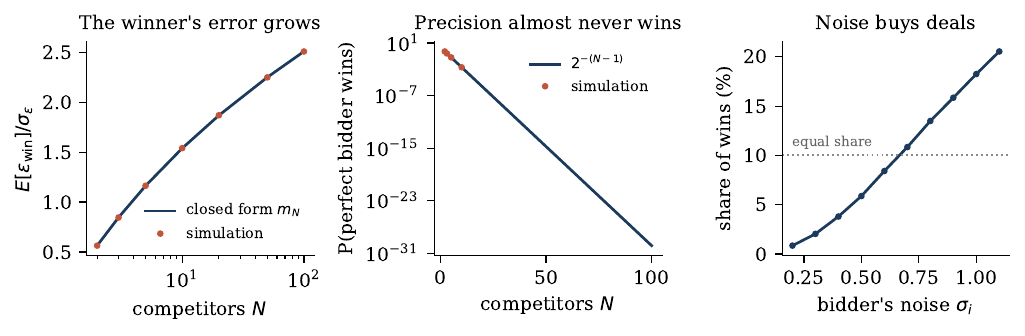}
\caption{\textbf{The \gate{max} gate.} Left: the winner's expected error grows like $m_N$. Center: a
bidder with a perfect estimate wins with probability $2^{-(N-1)}$; past $N\!=\!10$ this is below Monte
Carlo resolution and only the closed form is drawn. Right: ten bidders differing \emph{only} in noise.
The share of wins rises with a bidder's own noise, and so does the winner's overpayment, from $0.30$
to $1.41$ log-odds.}
\label{fig:max}
\end{figure}

\paragraph{The perfect bidder and $m_N$.} Under a common value with shared biases every bid is
$V+\varepsilon_i$, so $\arg\max_i \hat y_i=\arg\max_i\varepsilon_i$ (Eq.~\ref{eq:identity}). A bidder
with $\varepsilon_1=0$ wins if and only if $\max_{i\ge2}\varepsilon_i<0$; with independent errors
symmetric about zero, $\Pr(\text{win})=2^{-(N-1)}$, which is $1/512$ at $N=10$. The winner's expected
error is $\sigma_\varepsilon m_N$ with $m_N=N\!\int z\,\phi(z)\Phi(z)^{N-1}dz$, integrated
numerically; $m_{10}=1.539$ and $m_{50}=2.249$ agree with simulation to within $0.1\%$. Among ten
bidders whose noise $\sigma_i$ runs from $0.2$ to $1.1$, the most precise wins $0.85\%$ of contests
and the noisiest $20.5\%$, while the winner's overpayment rises with its noise, from $0.30$ to $1.41$
log-odds (Figure~\ref{fig:max}). If bidder $i$ also carries its own bias $\beta_i^\top x$, the winner
maximizes $\beta_i^\top x+\varepsilon_i$ and is selected on bias as well. Pooling before the contest
weakens the selection: if each bidder averages $k$ independent estimates, the overestimate falls to
$\sigma_\varepsilon m_N/\sqrt{k}$.

\paragraph{The selected item when qualities differ.} Let true quality $L\sim\mathcal N(\mu,t)$ and
score $y=L+e$ with error variance $V$, so $\sigma_y^2=t+V$, $\lambda=t/(t+V)$ and
$\E[L\mid y]=\mu+\lambda(y-\mu)$. The item with the highest of $N$ scores is overestimated by
$(1-\lambda)\E[y_{\max}-\mu]=(1-\lambda)\sigma_y m_N$. In terms of the error variance $V$
this is $\sqrt{V(1-\lambda)}\,m_N$, since $(1-\lambda)\sigma_y=\sqrt{V}\sqrt{V/(t+V)}$. At
$\lambda\to0$ every item is equally good and this is Eq.~\eqref{eq:identity}'s
$\sigma_\varepsilon m_N$. Since $V$ falls with the size of a test set, a large holdout keeps the
overestimate small.

\paragraph{The jury-scored leaderboard.} Let $N$ models of equal true quality each be scored by the
mean of $k$ judges. Judge $j$'s score of model $i$ is $\delta_{ij}+\varepsilon_{ij}$, where
$\delta_{ij}$, the judge's whole bias toward model $i$, has scale $\sigma_\beta$ and correlation
$\rho_\beta$ across judges for the same model and is independent across models, and
$\varepsilon_{ij}$ is independent noise with scale $\sigma_\varepsilon$. A part that every judge
holds toward model $i$ enters $\rho_\beta$, so $\sigma_\beta^2\rho_\beta$ is the common part and the
panel part together. A bias $c_j$ that judge $j$ gives every model equally adds the same constant to
every model's mean and cannot change which is highest. Model $i$'s averaged error is Gaussian with variance
$\sigma_\beta^2[\rho_\beta+(1-\rho_\beta)/k]+\sigma_\varepsilon^2/k$ by Eq.~\eqref{eq:avg},
independent across models, so the highest of $N$ averaged scores overestimates its model by $m_N$
times the square root of that variance, which is Eq.~\eqref{eq:jury}. As $k\to\infty$ the variance
falls to $\sigma_\beta^2\rho_\beta$, giving the floor $m_N\sigma_\beta\sqrt{\rho_\beta}$. At $N=20$,
$m_{20}=1.867$; with $\sigma_\beta=0.5$ and $\sigma_\varepsilon=1$ the overestimate is $2.09$ for one
judge, and for ten judges $0.66$ at $\rho_\beta=0$, $0.82$ at $\rho_\beta=0.3$ and $1.07$ at
$\rho_\beta=0.9$; the floors are $0.51$ at $\rho_\beta=0.3$ and $0.89$ at $\rho_\beta=0.9$. A
simulation with $20{,}000$ replications per setting agrees to the third decimal. With models of
unequal true quality, of variance $t$ independent of the errors, the averaged error variance plays the
part of $V$ in the previous paragraph, and the overestimate is multiplied by
$\sqrt{1-\lambda}=\sqrt{V/(t+V)}$.

\begin{figure}[h]
\centering
\includegraphics[width=0.86\linewidth]{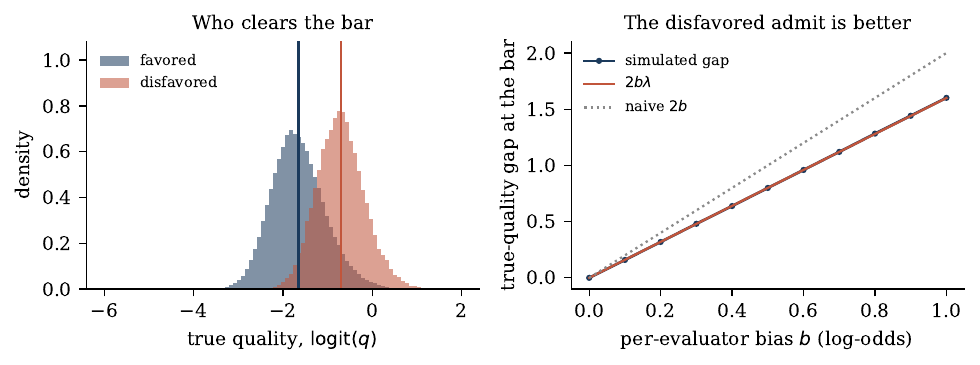}
\caption{\textbf{The \gate{threshold} gate.} Left: true quality of admitted candidates by group at a
gate that admits the top $10\%$ with $b=0.7$; vertical lines are group means. Right: the gap in true
quality at the bar tracks $2b\lambda$, not $2b$.}
\label{fig:thresh}
\end{figure}

\paragraph{The gap $2b\lambda$.} Let $L=\lgt(q)\sim\mathcal{N}(\mu_g,\sigma_T^2)$ for group $g$,
$\varepsilon\sim\mathcal{N}(0,\sigma_\varepsilon^2)$, and perceived score $y=L+c+\varepsilon$ with
$c=+b$ for the favored group and $-b$ for the disfavored. $(L,y)$ is jointly Gaussian, so
$\E[L\mid y]=\mu_g+\lambda(y-c-\mu_g)$. At the common bar $y=\tau$, the disfavored minus favored
conditional mean is $2b\lambda+(1-\lambda)(\mu_D-\mu_F)$, which is $2b\lambda$ when the groups' means
are equal. Because the conditional mean is linear in $y$, the simulation estimates it by a per-group
least-squares fit of $L$ on $y$ evaluated at $\tau$: $1.121$ against Eq.~\eqref{eq:gap}'s $1.120$. In
that simulation the median success probability is $5\%$ (the mean is $7.27\%$), the gate admits the
top $10\%$, $b=0.7$ and $\lambda=0.8$; the admitted pool is $92.1\%$ favored, and the disfavored admits'
mean true success probability is $0.341$ against the favored admits' $0.177$
(Figure~\ref{fig:thresh}).

\paragraph{Regression after selection.} From the same conditional mean with $\mu_g=\mu$, a selectee
scored $y$ falls short of its score by $y-\E[L\mid y]=(1-\lambda)(y-\mu)+\lambda c$. The shortfall
grows as $\lambda$ falls, that is, as transient factors weigh more, and the favored group ($c>0$)
regresses more.

\paragraph{Infra-marginality.} With no bias and group-specific quality
$L\sim\mathcal N(\mu_g,\sigma_{T,g}^2)$, the mean quality of a group's admits at a bar $\tau$ is
$\mu_g+(\sigma_{T,g}^2/\sigma_{y,g})\,h(z_g)$, with $\sigma_{y,g}^2=\sigma_{T,g}^2+\sigma_\varepsilon^2$,
$z_g=(\tau-\mu_g)/\sigma_{y,g}$ and $h(z)=\phi(z)/(1-\Phi(z))$. Admits' averages then differ by group
with no bias at all.

\paragraph{Berkson.} Selecting $y>\tau$ with $y=L+\varepsilon$ Gaussian and $z=(\tau-\mu_y)/\sigma_y$
gives $\mathrm{Cov}(L,\varepsilon\mid\text{admitted})=-(\sigma_T^2\sigma_\varepsilon^2/\sigma_y^2)\,
h(z)(h(z)-z)<0$, since $1-h(h-z)$, the variance of a truncated standard normal, lies in $(0,1)$. The
same algebra with the binary feature gives a negative association of quality and the favored feature
among admits: the outcome-test result is the same selection effect.

\paragraph{Group-dependent noise at a raw-score bar.} With equal $\mu$ and $\sigma_T$ and
group-specific $\sigma_{\varepsilon,g}$, the admit rate $1-\Phi((\tau-\mu)/\sigma_{y,g})$ rises with
$\sigma_{\varepsilon,g}$ whenever $\tau>\mu$, while the quality at the bar $\mu+\lambda_g(\tau-\mu)$
and the admits' mean $\mu+(\sigma_T^2/\sigma_{y,g})h(z_g)$ both fall with it. A Bayesian evaluator
that shrinks each score toward the mean under-admits the noisier group at a high bar instead.

\paragraph{What noise audits and consistency experiments see of a bias.} Two evaluators judging the
same case differ by $D=(\beta_i-\beta_j)^\top x+(\varepsilon_i-\varepsilon_j)=(\delta_i-\delta_j)^\top
x+(\varepsilon_i-\varepsilon_j)$, so the common part $\bar\beta$ cancels. With an isotropic covariance
of the varying bias (equally variable along every feature direction),
$\Var(D)=2(1-\rho_\beta)\sigma_\beta^2\|x\|^2+2\sigma_\varepsilon^2$: the panel part
$\rho_\beta\sigma_\beta^2\|x\|^2$ never appears. A consistency experiment compares two panels, $A$ and
$B$, of $k$ reviewers each, staffed independently. Write each panel's mean error, across candidates,
as the sum of three parts: a part both panels hold, of variance $S$ (the common part, and any bias
held across the field both are drawn from); the panel part, held within one panel only, of variance
$P$; and the reviewers' own parts and noise, of variance $I/k$. The panels' covariance is then
$\sigma_T^2+S$ and $\mathrm{corr}(A,B)=(\sigma_T^2+S)/(\sigma_T^2+S+P+I/k)$, while the validity of
either panel is $\sigma_T/\sqrt{\sigma_T^2+S+P+I/k}$. The panel part $P$ lowers agreement, as
noise does, and the experiment cannot tell the two apart; the part both panels hold, $S$, raises agreement and lowers validity, so high agreement is compatible with a large shared bias.
This is the reliability-versus-validity point of measurement theory, in which rater variance shared
across a design enters the universe score~\citep{brennan2001generalizability}.

\paragraph{Calibrating the noise audit.} The audit's pair measure~\citep{kahneman2021}, $|a-b|$ over
the pair's mean, equals $2\tanh(|D|/2)$ for log premiums differing by $D$. For Gaussian $D$, the
median of $|D|$ is $0.6745\,\mathrm{SD}(D)$, and $\mathrm{SD}(D)$ is $\sqrt2$ times the per-judge
standard deviation of the own error. A median of $55\%$ gives a per-judge log-SD of $0.59$, $43\%$
gives $0.46$, and the executives' expected $10\%$ gives $0.105$. By this calculation, which is ours
and not the audit's, the executives underestimated the noise variance $31.8$-fold. Premiums are not
log-odds, so the calibration is illustrative.

\paragraph{The crowd within.} Split a person's own error into a bias of variance $s_b^2$, independent
across people, and occasion noise of variance $s_e^2$, with $w=s_e^2/(s_b^2+s_e^2)$. Two occasions of
one person average to variance $s_b^2+s_e^2/2$, and $n$ people to $(s_b^2+s_e^2)/n$, so two own
guesses are worth $n_{\mathrm{eff}}=1/(1-w/2)$ people. \citet{vul2008} report $n_{\mathrm{eff}}$
directly: the number of different people whose averaged guesses would match the error of two own
guesses, read off the hyperbola fitted to averages across people (their $\lambda$ is
$n_{\mathrm{eff}}-1$). The hyperbola's asymptote, the bias common to the population, is the same for
both kinds of average and does not enter $n_{\mathrm{eff}}$. The measured $1.11$ gives $w\approx0.20$,
and $1.32$ gives $w\approx0.49$. Guesses made at once share noise, so these are lower bounds on $w$.

\paragraph{Bootstrapping.} Assume that a model of the judge reproduces the judge's view,
$\lgt(q)+\beta^\top x$, exactly and drops $\varepsilon$. Validity then rises from
$\sigma_T/\sqrt{\sigma_T^2+B+\sigma_\varepsilon^2}$ to $\sigma_T/\sqrt{\sigma_T^2+B}$, with $B$ the
bias variance. This is Eq.~\eqref{eq:avg} within one judge, $\rho_\beta=1$, $k\to\infty$. The
assumption carries the result: a lens-model model of the judge is fitted to
cues~\citep{karelaia2008}, and Eq.~\eqref{eq:model}'s judge sees true quality directly, so the
catalogue counts this row as mapped.

\paragraph{Swap-averaging.} A judge with position bias $\beta$ on $x=\pm1$ scores
$L+\beta x+\varepsilon$; swapping the order flips $x$, so the average of the two calls carries no
position term. A bias with no sign flip, such as a preference for length, is unchanged by any number
of calls: the floor of Eq.~\eqref{eq:avg} at $\rho_\beta=1$.

\paragraph{Committees' overlap and disagreement.} With two committees' scores bivariate normal at
correlation $r$ and each list cut at acceptance rate $a$, the share of one accepted list outside the
other is $1-\Pr(\text{both})/a$, and the share of papers on which they disagree is
$2(a-\Pr(\text{both}))$, which is $2a$ times the outside share: the two are one statistic. Each year
is cut at its own rate. In 2014 the duplicated papers were accepted at $43/166$ and $44/166$, a mean
of $26.2\%$; at $r=0.55$ the circuit gives $48.1\%$ outside (measured $49.5\%$) and $25.2\%$
disagreement (measured $25.9\%$, $43$ of $166$ papers). In 2021 the rate was $22.7\%$; the score
correlations of $0.575$ and $0.586$, for the initial and the final scores, give $49.6\%$ and $48.9\%$
outside (measured $50.6\%$) and $22.5\%$ and $22.2\%$ disagreement (measured $23.0\%$). The binomial
standard error of a measured outside share is about $3.5$ points in 2021, with about $200$ accepts per
committee, and about $7.6$ points in 2014, with about $43$. The 2014 correlation has its own
uncertainty, $0.55\pm0.065$, and across it the predicted outside share runs from $51.7\%$ at
$r=0.485$ to $44.4\%$ at $r=0.615$, so the comparison cannot discriminate between values of $r$ over
that range. No bias term enters: the comparison tests the \gate{average} $\to$ \gate{threshold}
structure, not the persistent bias. The 2014 correlation is between mean calibrated scores and the
2021 correlations between mean uncalibrated scores, and real committees do not cut mean scores
mechanically, so the comparison checks the structure, not the committees' procedure. At $r=0.55$ and
$a=10,15,22.7,30,40,50\%$ the outside shares are $64.4$, $58.5$, $51.1$, $45.1$, $38.0$ and $31.5\%$
(at $22.7\%$ the disagreement is $23.2\%$); this decline holds for any bivariate-normal model with a
hard cut. For two median cuts, disagreement is $\tfrac12-\arcsin(r)/\pi$: $29.5$, $28.7$ and $27.1\%$
at $r=0.60$, $0.62$ and $0.66$, against about $25\%$ of decisions reported reversed in the NSF
re-review~\citep{cole1981}; the definitions may differ, so the comparison is approximate.

\paragraph{Panel agreement from one reviewer's reliability.} Eq.~\eqref{eq:avg} with
$\rho_\varepsilon=0$, applied to reliability, gives the reliability of a mean of $k$ ratings,
$kr/(1+(k-1)r)$. At $r=0.34$ it gives $0.61$ for $k=3$ and $0.67$ for $k=4$, the predicted
correlation between two independent panels of that size. The value $0.34$ is a pooled estimate from a
meta-analysis that combined several reliability measures~\citep{bornmann2010}, so it serves only as an
illustrative single-reviewer value. The sources do not state how many reviewers each paper had in each
year, so both predictions are reported. Both lie above the correlations measured at NeurIPS ($0.55$,
$0.575$, $0.586$); the value for three reviewers lies within the $0.60$--$0.66$ of the NSF
re-review~\citep[p.~882]{cole1981}. Reading the reliability of a mean of ratings as
Eq.~\eqref{eq:avg} with $\rho_\varepsilon=0$ is an informal identification of rating error with
occasion noise.

\paragraph{Agreement between two fresh reviewers after a screen.}
Let $L$ be latent proposal quality, $S=L+e_0$ the screening score, and $R_i=L+e_i$ the
scores of two new reviewers. Let $L,e_0,e_1,e_2$ be independent centered Gaussians, with
$\Var(L)=t>0$ and $\Var(e_1)=\Var(e_2)=s^2>0$. Retain the top fraction $p$ on $S$.
With $a=\Phi^{-1}(1-p)$, $d=\phi(a)/p$ and $u^2=1-d(d-a)$, the truncated screen has variance
$u^2\Var(S)$. Regressing $L$ on $S$ leaves independent residual variance $t(1-\lambda_S)$,
where $\lambda_S=t/\Var(S)$. The surviving between-proposal variance and the fresh reviewers'
correlation are therefore
\begin{equation}
 v=\Var(L\mid\mathrm{pass})=t[1-\lambda_S(1-u^2)],\qquad
 \operatorname{Corr}(R_1,R_2\mid\mathrm{pass})=\frac{v}{v+s^2}.
 \label{eq:fresh-review}
\end{equation}
The fresh reviewers share only $L$, so their conditional covariance is $v$. An informative
screen gives $v<t$, and agreement falls below its unrestricted value $r=t/(t+s^2)$.
For $r=0.34$ and a top-$10\%$ cut, agreement falls to $0.270$ if the screen is one equally
reliable review. If it averages three independent reviews, $\lambda_S=3r/(1+2r)$ and
agreement falls to $0.203$. These illustrate attenuation while leaving the two new reviewers'
noise variances fixed.

\paragraph{Interview validity at fixed panel size.}
Let two independent, centered, unit-variance cues $X_1,X_2$ predict performance
$L=X_1+X_2$. Interviewer $i$ gives score $Y_i=L+b_i(X_1-X_2)+\varepsilon_i$: the cue
weights are $1+b_i$ and $1-b_i$. The $b_i$ stay fixed across candidates and the occasion
errors are independent across interviewers and independent of the cues, with variance
$\sigma_\varepsilon^2$. For the same $k$ interviewers, a rubric prescribes the criterion
weights $(1,1)$, while unstructured scoring leaves the weights to each interviewer.
Since $X_1-X_2$ is uncorrelated with $L$ and both have variance $2$, the panel mean has
\begin{equation}
 \operatorname{Corr}(\bar Y,L)
 =\bigl[1+\bar b^2+\sigma_\varepsilon^2/(2k)\bigr]^{-1/2},
 \qquad \bar b=k^{-1}\textstyle\sum_i b_i .
 \label{eq:cue-validity}
\end{equation}
The rubric sets $\bar b=0$. For example, with two interviewers, deviations $0.2$ and $0.8$,
and zero occasion noise, validity falls from $1$ to $0.894$. Deviations of $0.2$ and $-0.2$
cancel exactly. If judges' deviations are independent draws with mean zero and variance
$\sigma_b^2$, the panel's mean squared deviation is $\E[\bar b^2]=\sigma_b^2/k$.
The loss comes from the cue-weight error left in the panel mean, and assumes
that the rubric weights the cues in line with the performance criterion.

\begin{figure}[h]
\centering
\includegraphics[width=0.86\linewidth]{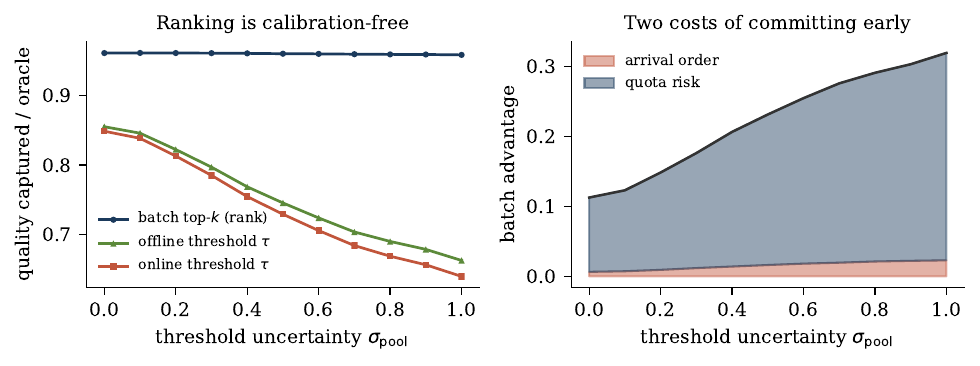}
\caption{\textbf{\gate{rank} versus \gate{threshold}.} Left: quality captured against an oracle,
versus uncertainty in the pool's mean; \gate{rank} (``batch top-$k$'' in the legend, keeping the top
$s=10$) is flat. Right: the advantage decomposed. Quota risk dominates, and is already large at a
perfectly centered threshold. Unfilled slots count as zero in the metric.}
\label{fig:rank}
\end{figure}

\paragraph{\gate{rank} versus \gate{threshold} under drift.} The simulation of \S\ref{sec:rank} draws
$n=100$ candidates for $s=10$ slots, with a pool mean that drifts unseen, and scores each rule by the
true quality its admits capture against an oracle that takes the truly best $s$. \gate{rank} captures
$0.96$ whatever the drift, flat to within $0.003$. An online threshold, committed in advance and
taking the first $s$ to clear it, slides from $0.85$ with no drift to $0.64$ at the largest. An offline
threshold, which sees the whole pool, shows where the loss comes from: at every drift, more than nine
tenths of it is quota risk, slots left unfilled. Even a perfectly centered bar fills only $88\%$ of the
slots, because the number of candidates who clear it varies from pool to pool, and a pool whose mean
drifts by one standard deviation of quality leaves the bar filling $67\%$. That share partly reflects
the metric, which scores an unfilled slot zero (Appendix~\ref{app:se} gives the decomposition;
Figure~\ref{fig:rank}).

\paragraph{Monoculture.} With $n$ firms' errors equicorrelated at $\rho$ through the one-factor
construction $v_j=\sqrt\rho\,s+\sqrt{1-\rho}\,u_j$, the probability that a candidate is rejected by all
$n$ firms is $\E_s[\Phi((\tau-\sqrt\rho\,s)/\sqrt{1-\rho})^n]$, a one-dimensional integral. Only the
bias part of $\rho$ persists when the same candidate applies again.

\clearpage
\section{Parameters of every experiment}
\label{app:params}

All quantities are in log-odds unless stated. The same values appear as named constants in the
code, which will be released. The main script draws each of its sections from its own random stream derived from one
master seed; the further runs named below (the bias-correlation sweep, the sampling floor, the price
of a panel, the executive gate's controls and crossing, the leaderboard check and the team's exact
computation) each use their own generator.

\begin{table}[h]
\centering
\small
\begin{tabular}{@{}lp{0.72\linewidth}@{}}
\toprule
\textbf{Section} & \textbf{Parameters} \\
\midrule
\S\ref{sec:2x2} three interviewers &
$k=3$; $b=0.6$; $\sigma_\varepsilon=1$; $\lgt(q)\sim\mathcal{N}(0,1)$; hire rate $0.10$;
$p\in\{0.50,0.35,0.20\}$; $2500$ employers $\times$ $500$ candidates per cell. Panel threshold:
pooled quantile of the panel mean. Hurdles: one per-stage bar found by bisection so that three-stage
survival equals the hire rate. Biases drawn once per employer: all $+b$ (a common part $\bar\beta=b$,
no varying part) or independent $\pm b$ (idiosyncratic signs, $\bar\beta=0$). Bias-correlation run
(Table~\ref{tab:rhosweep}): $\beta_i=\bar\beta+\delta_i$, $\delta_i$ Gaussian with $\sigma_\beta=0.6$,
equicorrelated at $\rho_\beta\in\{0,0.3,0.6,0.9,1\}$ by the one-factor construction;
$\bar\beta\in\{0,0.3\}$; $p=0.35$. Sampling-floor run (Table~\ref{tab:floor}): the same design at
$\bar\beta=0$, with $\sigma_\beta\in\{0,0.6\}$. Interviews per applicant: the hurdles' design at
$p=0.35$, both bias patterns. \\
\S\ref{sec:average} \gate{average} &
$\sigma=1$; $k=1,\dots,24$; $\rho\in\{0,0.3,0.6,0.9\}$; $200{,}000$ draws per $(k,\rho)$;
equicorrelation by a one-factor construction $v_j=\sqrt{\rho}\,s+\sqrt{1-\rho}\,u_j$. \\
\S\ref{sec:max} \gate{max} &
$m_N$: $400{,}000$ draws per $N\in\{2,3,5,10,20,50,100\}$. Perfect bidder: $4{,}000{,}000$ draws,
checked where the expected success count is at least $2000$. Different precision: ten bidders,
$\sigma_i=0.2,0.3,\dots,1.1$, common value $0$, $400{,}000$ auctions. \\
\S\ref{sec:threshold} \gate{threshold} &
$\lgt(q)\sim\mathcal{N}(\lgt 0.05,\,1)$, so the median success probability is $5\%$ and the mean
$7.27\%$; $\sigma_T=1$; $\sigma_\varepsilon=0.5$ ($\lambda=0.8$); $b=0.7$; groups equiprobable;
acceptance $0.10$; $4{,}000{,}000$ candidates. Sweep: $b\in\{0,0.1,\dots,1\}$ at $1{,}200{,}000$ each. \\
\S\ref{sec:rank} \gate{rank} vs \gate{threshold} &
$n=100$; $s=10$; $\sigma_T=1$; $\sigma_\varepsilon=0.5$; $\tau=\Phi^{-1}(0.9)\sqrt{\sigma_T^2
+\sigma_\varepsilon^2}$; pool shift $m\sim\mathcal{N}(0,\sigma_{\mathrm{pool}}^2)$,
$\sigma_{\mathrm{pool}}\in\{0,0.1,\dots,1\}$; $40{,}000$ cycles per $\sigma_{\mathrm{pool}}$; quality
captured $=\sum q$ of admits $/\sum q$ of the true top $s$. \\
\S\ref{sec:homophily} self-staffing team &
Exact computation: team $T=12$; $N=60$ applicants per opening, types equiprobable; true quality
$\mathcal N(0,1)$ plus the panel's noise of variance $1/T$, so $\sigma_w=\sqrt{1+1/12}$; hire = top of
the panel mean (\gate{rank}, $s=1$ hire); the hire replaces a uniformly random seat; $40$ replacements
from a coin flip per seat; $h\in[0,1.5]$; pool sweep $N\in\{10,20,30,60,120,250,500\}$. Simulation
check: $4000$ organizations, $\sigma_\varepsilon=1$ with panel noise $\sigma_\varepsilon/\sqrt{T}$,
$h\in\{0,0.3,0.6,1.0\}$; at $h=0$, five further seeds. \\
\S\ref{sec:exec} last decision &
$k=8$; $\sigma_\beta=0.5$; $\sigma_\varepsilon=0.7$ (panel noise $\sigma_\varepsilon/\sqrt{8}$, one
person's noise $\sigma_\varepsilon$); two panel stages each advancing $40\%$; the final evaluator
advances $40\%$ of survivors on its own score; $4000$ candidates per firm, tracked attribute
$z\sim\mathrm{Bernoulli}(0.5)$; panels' biases equicorrelated at $\rho_\beta\in\{0,0.2,0.5,0.8\}$
within a panel, independent across panels; the executive's bias $\mathcal{N}(0,\sigma_\beta^2)$.
Crossing: $2{,}000$ firms per $\rho_\beta$, the final evaluator's bias (a fresh panel of eight or one
person) crossed with its noise (averaged over eight or one person's), on the same survivors. Controls
(Table~\ref{tab:exec}): $600$ firms per $\rho_\beta$, each control a third $40\%$ cut: C1 on the
second panel's score; C2 a new panel of eight at the same $\rho_\beta$; C3 a new panel of seven at that
$\rho_\beta$ plus one member with an independent bias of scale $\sigma_\beta$. \\
\S\ref{sec:jury} leaderboard &
$N=20$ models of equal quality; $\sigma_\beta=0.5$; $\sigma_\varepsilon=1$; $k\in\{1,3,10,100\}$;
$\rho_\beta\in\{0,0.3,0.9\}$; $20{,}000$ replications per cell for the simulation check. \\
Appendix~\ref{app:static} fixed series &
six \gate{threshold} stages, each advancing half the survivors; $b\in\{0.1,0.2,0.4\}$ for the
in-group; $\sigma_\varepsilon=0.6$; $\sigma_T=1$; groups equiprobable; $1{,}500{,}000$ candidates. \\
\bottomrule
\end{tabular}
\end{table}

\section{Monte Carlo standard errors}
\label{app:se}

Standard errors are computed from the spread of the same run: binomial for shares, the per-draw
standard deviation over $\sqrt{n}$ for means, and $s/\sqrt{2(n-1)}$ for a standard deviation $s$ over
$n$ organizations. For the shares and means they sit below the body's last digit. For spreads across
organizations they are $0.1$ to $0.6$ points: $0.10$ to $0.31$ in the executive crossing over $2{,}000$
firms, and $0.2$ to $0.6$ in the controls over $600$. The executive crossing's hires' mean success has
a standard error below $0.02$ points in every arm, and the paired gap between a fresh panel and one
executive, $2.0$ to $2.3$ points across $\rho_\beta$, has a standard error below $0.02$. The standard
errors of the three-interviewer run treat hires as independent; with biases drawn once per employer,
the idiosyncratic-bias cells' errors are larger, at least $0.006$ for the $1.07\times$ cell, and no
cluster-robust errors were computed. Even so, the idiosyncratic-bias values, $1.07$ in the panel and
$0.96$ in the hurdles, differ from $1$ and from each other by more than these errors. In the
bias-correlation run the per-employer skew over $2{,}500$ employers has a standard error of about
$0.005$, and pooled over-representation at most $0.007$, treating hires as independent and so
understated. The team's values in \S\ref{sec:homophily} are exact; the simulation that checks them is
reported in Appendix~\ref{app:static}.

\begin{table}[h]
\caption{Monte Carlo standard errors of the main simulated values.}
\label{tab:se}
\centering
\small
\begin{tabular}{@{}llll@{}}
\toprule
\textbf{Quantity} & \textbf{Value} & \textbf{SE} & $n$ \\
\midrule
win share, $\sigma_i=0.2$ & $0.851\%$ & $0.015\%$ & $400{,}000$ auctions \\
win share, $\sigma_i=1.1$ & $20.55\%$ & $0.06\%$ & $400{,}000$ auctions \\
mean true $q$, favored admits & $0.1768$ & $0.0002$ & $368{,}500$ admits \\
mean true $q$, disfavored admits & $0.3412$ & $0.0007$ & $31{,}500$ admits \\
favored share of admits & $0.9213$ & $0.0004$ & $400{,}000$ admits \\
spread, fresh panel, $\rho_\beta=0$ & $8.37$\,pp & $0.13$\,pp & $2000$ firms \\
spread, one executive, $\rho_\beta=0$ & $13.92$\,pp & $0.22$\,pp & $2000$ firms \\
spread, fresh panel, $\rho_\beta=0.8$ & $17.54$\,pp & $0.28$\,pp & $2000$ firms \\
spread, one executive, $\rho_\beta=0.8$ & $15.01$\,pp & $0.24$\,pp & $2000$ firms \\
hires' mean success, fresh panel, $\rho_\beta=0.8$ & $85.88\%$ & $0.016\%$ & $2000$ firms \\
hires' mean success, one executive, $\rho_\beta=0.8$ & $83.87\%$ & $0.015\%$ & $2000$ firms \\
\gate{rank} quality captured, $\sigma_{\mathrm{pool}}=0$ & $0.9612$ & $0.0001$ & $40{,}000$ cycles \\
online \gate{threshold}, $\sigma_{\mathrm{pool}}=0$ & $0.8486$ & $0.0008$ & $40{,}000$ cycles \\
online \gate{threshold}, $\sigma_{\mathrm{pool}}=1$ & $0.6395$ & $0.0018$ & $40{,}000$ cycles \\
mean true $q$ of hires (\S\ref{sec:2x2}, any cell) & $0.78$--$0.80$ & $0.0003$ & $125{,}000$ hires \\
over-representation (\S\ref{sec:2x2}, any cell; hires as independent) & $0.94$--$1.85$ & $\le 0.007$ & $125{,}000$ hires \\
\bottomrule
\end{tabular}
\end{table}

The decomposition of \S\ref{sec:rank}'s gap: of the $0.113$ between \gate{rank} and the online
threshold at zero drift, $0.106$ is quota risk; at $\sigma_{\mathrm{pool}}=1$ the figures are $0.296$ of
$0.319$.

\clearpage
\section{The three-interviewer circuit in full}
\label{app:perorg}

\paragraph{Design.} Each interviewer is the evaluator of Eq.~\eqref{eq:model} with $x$ reduced to one
binary attribute $z$, coded $0$ or $1$ and unrelated to quality, which makes the bias a single number.
In the first runs an interviewer adds $+b$ or $-b$, one sign per interviewer, to every judgment of a
candidate with $z=1$ and nothing to a candidate with $z=0$, while the noise is drawn afresh for each
judgment. With $b=0.6$ and $\sigma_\varepsilon=1$ against a spread of $1$ in true quality, the noise in
a single judgment is as large as the applicants' spread, and the bias shifts a judgment by somewhat
more than half of it. Under a common part, every interviewer at every employer adds $+b$ to the
tracked group: $\bar\beta=b$, which no average removes. Under idiosyncratic biases each sign is drawn
at even odds, independently of the other interviewers and of every other employer, so $\bar\beta=0$,
the varying part is uncorrelated, and no group is favored across employers. These two conditions
differ in the common part as well as in the correlation of the varying part; the bias-correlation run
below separates the two. The panel's one threshold on its mean and the hurdles' one threshold shared
by their three gates are each set once for all employers together, so that luck in one employer's
applicants cannot move them. The designs share one hire rate, one tracked group, one size of bias and
one level of noise, so any difference in their hires is the effect of a design choice.

\paragraph{The bias correlation at a fixed common part.} A second run separates the two parts
(Table~\ref{tab:rhosweep}): each interviewer's bias is $\beta_i=\bar\beta+\delta_i$, with the
$\delta_i$ Gaussian, of scale $\sigma_\beta=0.6=b$ so that the varying part matches the $\pm b$ signs
in variance, and equicorrelated at $\rho_\beta$ within an employer; the tracked group is $35\%$ of
applicants and the hire rate $10\%$. The $\rho_\beta=0$ row reproduces the idiosyncratic cells closely
($1.07$ and $0.31$ in the panel; $0.97$ and $0.30$ in the hurdles, against $0.96$ and $0.30$). At
$\bar\beta=0$, raising $\rho_\beta$ from $0$ to $1$ raises the per-employer skew by about $45\%$ in both
designs, lowers the hires' quality slightly and moves pooled over-representation less. At
$\rho_\beta=0$, raising $\bar\beta$ from $0$ to $0.3$ moves pooled over-representation from $1.07$ to
$1.36$ in the panel. The common part moves who is hired across all employers, and the correlation of
the varying part moves the skew inside them: the two parts of the bias are separable in their effects.
In every row the panel's and the hurdles' skews are close, so the combining rule cannot reduce the
skew inside an employer; lowering the correlation of the interviewers' biases does. That a bias common
to every evaluator moves pooled representation through selection stages is known: it is test bias in
the sense of \citet{cleary1968test}, carried through multistage selection~\citep{decorte2006}.

\begin{table}[h]
\caption{The bias correlation at a fixed common part: pooled over-representation, per-employer skew,
and the hires' mean true success probability, for the panel and the hurdles. $2{,}500$ employers
$\times$ $500$ candidates per row; tracked group $35\%$; hire rate $10\%$; $\sigma_\beta=0.6$,
$\sigma_\varepsilon=1$.}
\label{tab:rhosweep}
\centering
\small
\begin{tabular}{@{}ll|ccc|ccc@{}}
\toprule
 & & \multicolumn{3}{c|}{panel} & \multicolumn{3}{c}{hurdles} \\
$\bar\beta$ & $\rho_\beta$ & over-rep. & skew & quality & over-rep. & skew & quality \\
\midrule
$0$ & $0.0$ & $1.07$ & $0.31$ & $0.799$ & $0.97$ & $0.30$ & $0.783$ \\
$0$ & $0.3$ & $1.10$ & $0.37$ & $0.797$ & $1.02$ & $0.35$ & $0.782$ \\
$0$ & $0.6$ & $1.13$ & $0.40$ & $0.795$ & $1.07$ & $0.38$ & $0.780$ \\
$0$ & $0.9$ & $1.18$ & $0.44$ & $0.791$ & $1.15$ & $0.42$ & $0.777$ \\
$0$ & $1.0$ & $1.17$ & $0.45$ & $0.791$ & $1.15$ & $0.43$ & $0.777$ \\
$0.3$ & $0.0$ & $1.36$ & $0.40$ & $0.795$ & $1.24$ & $0.34$ & $0.780$ \\
$0.3$ & $0.3$ & $1.40$ & $0.45$ & $0.791$ & $1.30$ & $0.40$ & $0.778$ \\
$0.3$ & $0.6$ & $1.44$ & $0.49$ & $0.788$ & $1.36$ & $0.44$ & $0.775$ \\
$0.3$ & $0.9$ & $1.47$ & $0.53$ & $0.785$ & $1.42$ & $0.49$ & $0.772$ \\
$0.3$ & $1.0$ & $1.50$ & $0.55$ & $0.782$ & $1.46$ & $0.52$ & $0.770$ \\
\bottomrule
\end{tabular}
\end{table}

\paragraph{The sampling floor and the relative spread across employers.} With about fifty hires per
employer, the skew $|\text{share}_{\mathrm{employer}}/p-1|$ has a floor from binomial sampling alone:
$0.16$ in the panel and $0.15$ in the hurdles with no bias at all (Table~\ref{tab:floor}). The body
therefore reports a corrected measure, the \emph{relative spread across employers}: the standard deviation across
employers of the tracked group's share of hires, with the mean binomial variance subtracted, divided
by $p$. It is near zero with no bias ($0.02$ and $0.00$). Once sampling is removed, raising
$\rho_\beta$ from $0$ to $1$ widens the spread by about $55\%$ ($0.32$ to $0.50$ in the panel, $0.31$
to $0.48$ in the hurdles), more than the $45\%$ the raw skew suggests. The floor also grows as the
tracked group's expected number of hires falls, so part of the rise of the skew as $p$ falls in
Table~\ref{tab:perorg} is sampling rather than bias.

\begin{table}[h]
\caption{The skew's sampling floor and the relative spread across employers ($\bar\beta=0$; tracked
group $35\%$; hire rate $10\%$; $2{,}500$ employers $\times$ $500$ candidates per row; about fifty
hires per employer).}
\label{tab:floor}
\centering
\small
\begin{tabular}{@{}ll|cc|cc@{}}
\toprule
 & & \multicolumn{2}{c|}{panel} & \multicolumn{2}{c}{hurdles} \\
$\sigma_\beta$ & $\rho_\beta$ & skew & relative spread & skew & relative spread \\
\midrule
$0$ & -- & $0.16$ & $0.02$ & $0.15$ & $0.00$ \\
$0.6$ & $0.0$ & $0.30$ & $0.32$ & $0.30$ & $0.31$ \\
$0.6$ & $0.3$ & $0.37$ & $0.40$ & $0.35$ & $0.38$ \\
$0.6$ & $0.6$ & $0.40$ & $0.45$ & $0.38$ & $0.42$ \\
$0.6$ & $0.9$ & $0.45$ & $0.51$ & $0.43$ & $0.48$ \\
$0.6$ & $1.0$ & $0.45$ & $0.50$ & $0.43$ & $0.48$ \\
\bottomrule
\end{tabular}
\end{table}

\paragraph{The price of a panel.} The applicants' mean true quality is $50\%$ by symmetry, and the
hires' is near $80\%$ in every design. The hurdles' hires fall below the panel's by $1.2$ to $1.7$
percentage points across the runs: $1.4$ with the common part $\bar\beta=b$ and $1.7$ with
idiosyncratic biases in the first runs at $p=35\%$ (Table~\ref{tab:perorg}), and $1.2$ to $1.6$ in the
bias-correlation run (Table~\ref{tab:rhosweep}). The hurdles hire in effect on the lowest of three
noisy judgments, while the panel's mean carries a third of the noise variance; this is the mean of
three draws against their minimum, not the predictor composite of multistage selection
studies~\citep{decorte2006}. The panel pays for this in interviews. It interviews every applicant
three times, while the hurdles stop at the first veto: with the per-stage bar of these runs, they use
$1.445$ interviews per applicant under the common part and $1.466$ under idiosyncratic biases. A bias
costs true quality as well, because at a fixed hire rate it admits candidates of the favored group
with less true quality than the bar demands of the other (\S\ref{sec:threshold}); the common part
costs about half a point more than idiosyncratic biases in the panel and under a third of a point more
in the hurdles.

\paragraph{Who is hired, pooled.} Over-representation is the tracked group's share of the hires divided
by its share of applicants $p$, all employers counted together; $1$ is what an unbiased circuit gives,
because $z$ is unrelated to quality. With three interviewers and idiosyncratic signs, one employer in
four has all three leaning the same way and is as skewed as under the common part; the other three in
four have a net bias of $b/3$. Idiosyncratic biases accordingly about halve the skew a typical employer
experiences, $0.62$ to $0.31$ in the panel and $0.58$ to $0.30$ in the hurdles at $p=35\%$, and remove
it only in aggregate; about half of $0.31$ is the sampling floor above. Pooled over-representation under
idiosyncratic biases sits near parity, $1.07$ in the panel and $0.96$ in the hurdles, but the two differ
from $1$ and from each other by more than their standard errors. The asymmetry is the combining
rule's: in a series of hurdles a candidate must clear every interviewer, so the most hostile one binds,
while the panel's cut on the upper tail of the mean is convex in the shift. Under idiosyncratic biases
the combining rule therefore moves pooled over-representation by about $0.1$, more than the $0.04$
($1.62$ against $1.58$) by which it moves it under the common part.

\paragraph{Scarcity.} Over-representation is a ratio to $p$, so the same bias reads differently as $p$
changes. In the panel with the common part it rises from $1.44$ at $p=50\%$ to $1.85$ at $p=20\%$;
under idiosyncratic biases it stays near parity at every $p$ (Table~\ref{tab:perorg}). A second
measure, the gain, does not divide by $p$: it is the tracked group's share of hires minus its share of
applicants, in percentage points. In the panel with the common part the gain moves the other way, from
$22$ points at $p=50\%$ and $35\%$ to $17$ at $20\%$. Over-representation equals $1+\text{gain}/p$, and
it rises because $p$ falls by more than half while the gain falls by less than a quarter. Read alone,
over-representation shows a larger effect of the same bias on a scarcer group, although the bias moves
fewer hires.

\begin{table}[h]
\caption{The first runs of the three-interviewer circuit. ``Common'' is the common part
$\bar\beta=b$, ``idiosyncratic'' independent signs. The last column is the hires' mean true success
probability.}
\label{tab:perorg}
\centering
\small
\begin{tabular}{@{}llrrr@{}}
\toprule
$p$ & rule, bias & pooled over-rep. & per-employer skew & mean true quality \\
\midrule
$0.50$ & panel, common & $1.44\times$ & $0.44$ & $0.795$ \\
$0.50$ & panel, idiosyncratic & $1.05\times$ & $0.24$ & $0.798$ \\
$0.50$ & hurdles, common & $1.41\times$ & $0.41$ & $0.780$ \\
$0.50$ & hurdles, idiosyncratic & $0.97\times$ & $0.23$ & $0.781$ \\
$0.35$ & panel, common & $1.62\times$ & $0.62$ & $0.795$ \\
$0.35$ & panel, idiosyncratic & $1.07\times$ & $0.31$ & $0.800$ \\
$0.35$ & hurdles, common & $1.58\times$ & $0.58$ & $0.781$ \\
$0.35$ & hurdles, idiosyncratic & $0.96\times$ & $0.30$ & $0.783$ \\
$0.20$ & panel, common & $1.85\times$ & $0.85$ & $0.797$ \\
$0.20$ & panel, idiosyncratic & $1.07\times$ & $0.42$ & $0.801$ \\
$0.20$ & hurdles, common & $1.79\times$ & $0.79$ & $0.782$ \\
$0.20$ & hurdles, idiosyncratic & $0.94\times$ & $0.40$ & $0.785$ \\
\bottomrule
\end{tabular}
\end{table}

\clearpage
\section{Feedback and the fixed series}
\label{app:static}

\paragraph{The team's exact computation.} Let $a$ be the number of type-A members in a team of $T=12$,
and $f=a/T$. When the panel averages its members' judgments, an A applicant gains $hf$ and a B
applicant $h(1-f)$, a net pull of $d=h(2f-1)$ toward the current majority. Each applicant's averaged
judgment is true quality, of variance $1$, plus the panel's noise, of variance $1/T$, so judgments
within a type have spread $\sigma_w=\sqrt{1+1/T}$. With $N$ applicants of equiprobable type, the
chance that the top one is type A is
\[
H(f)=\int \tfrac{N}{2}\,\phi(z-d/\sigma_w)\left[\tfrac12\Phi(z)+\tfrac12\Phi(z-d/\sigma_w)\right]^{N-1}dz .
\]
The departing member is type A with chance $f$, so the team moves from $a$ to $a+1$ with chance
$(1-f)H(f)$ and to $a-1$ with chance $f(1-H(f))$, and its expected drift is $H(f)-f$. Starting from a
coin flip per seat, the distribution after $40$ replacements is the $13$-state chain's, computed
exactly. The expected drift vanishes at balance. Its slope there changes sign as follows. Since
$\partial H/\partial d=m_N/(4\sigma_w)$ at $d=0$ and $\partial d/\partial f=2h$, the slope
$\partial H/\partial f=2h\,m_N/(4\sigma_w)$ reaches $1$ at $h^*=2\sigma_w/m_N$, which is $0.898$ at
$N=60$. The crossing at $7$--$5$, where $H(7/12)=7/12$, is $0.905$, and the expected drift is toward
balance at every unbalanced composition for $h\le0.9$.

\paragraph{How long a majority lasts.}
Start at the first $9$--$3$ split and let $T_6$ count replacements through the next tie.
It includes time spent at $8$--$4$ or $7$--$5$ while the same type still leads. From the
transition probabilities above, form $Q$ by keeping the rows and columns for $a=7,\ldots,12$.
If $e_9$ denotes the coordinate vector for state $9$ in this ordering, then
\begin{equation}
 \Pr_9(T_6>m)=e_9^\top Q^m\mathbf 1,\qquad
 \operatorname{median}(T_6)=\min\{m\ge0:e_9^\top Q^m\mathbf 1\le1/2\}.
 \label{eq:team-return}
\end{equation}
At $60$ applicants per opening the medians are $12$, $22$, $40$, $56$ and $1{,}331$
replacements for $h=0,\,0.6,\,0.9,\,1,\,1.5$, respectively. Symmetry gives the same times
from $3$--$9$. At every finite $h$ both applicant types have a positive chance of winning,
so the finite chain returns to balance with probability one. The feedback changes how long
the early majority persists.

\begin{table}[h]
\caption{The self-staffing team, exact. Left: by own-type preference $h$ at $60$ applicants per
opening. Right: the threshold by applicants per opening (the crossing at $7$--$5$, with the closed
form at balance in brackets).}
\label{tab:team}
\centering
\small
\begin{tabular}{@{}lcc@{}}
\toprule
$h$ & 7--5 majority wins next hire & at or past $75/25$ after $40$ \\
\midrule
$0$ & $50.00\%$ & $14.60\%$ \\
$0.3$ & $52.78\%$ & $23.48\%$ \\
$0.6$ & $55.55\%$ & $38.11\%$ \\
$0.9$ & $58.29\%$ & $58.00\%$ \\
$1.0$ & $59.20\%$ & $64.77\%$ \\
$1.5$ & $63.63\%$ & $88.94\%$ \\
\bottomrule
\end{tabular}
\qquad
\begin{tabular}{@{}lc@{}}
\toprule
applicants & threshold $h^*$ \\
\midrule
$10$ & $1.36$ ($1.35$) \\
$20$ & $1.12$ ($1.12$) \\
$30$ & $1.03$ ($1.02$) \\
$60$ & $0.90$ ($0.90$) \\
$120$ & $0.82$ ($0.81$) \\
$250$ & $0.74$ ($0.74$) \\
$500$ & $0.69$ ($0.69$) \\
\bottomrule
\end{tabular}
\end{table}

\paragraph{The simulation as a check.} A simulation of the same loop over $4{,}000$ organizations gives
$13.7\%$ of teams at or past $75/25$ at $h=0$ and $65.0\%$ at $h=1$, draws around the exact $14.6\%$
and $64.8\%$. At $h=0$ the team's makeup is exactly Binomial$(12,\nicefrac12)$, with a mean deviation
from balance of $11.28$ points; the simulation gives $11.0$ (standard error $0.14$), and five further
seeds give $11.25$ to $11.65$ points and $14.5$ to $16.3\%$ past $75/25$, scattered about the exact
values. A hiring panel held at balance has $f=\nicefrac12$ and so $d=0$ at any $h$: it has no pull
toward either type.

\begin{figure}[h]
\centering
\includegraphics[width=\linewidth]{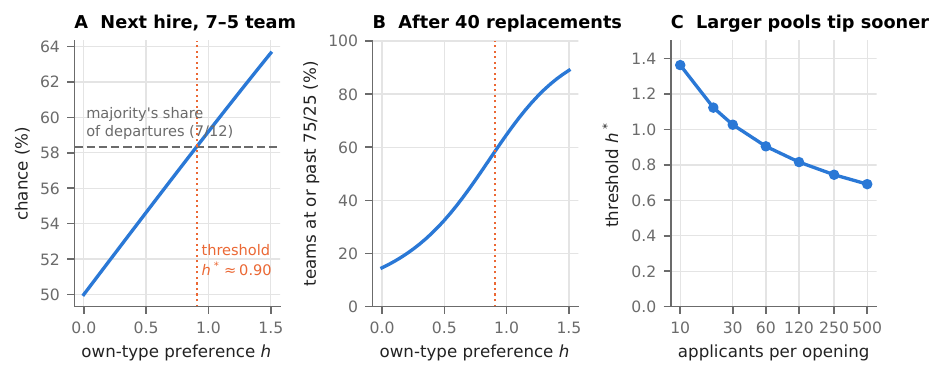}
\caption{\textbf{The self-staffing team, exact.} A: the chance that a $7$--$5$ majority wins the next
hire rises with own-type preference $h$ and crosses the majority's share of departures ($7/12$) at the
threshold $h^*\approx0.90$. B: the share of teams at or past $75/25$ after $40$ replacements rises with
$h$ on both sides of the threshold. C: the threshold falls as the number of applicants per opening
grows.}
\label{fig:team}
\end{figure}

\begin{figure}[h]
\centering
\includegraphics[width=\linewidth]{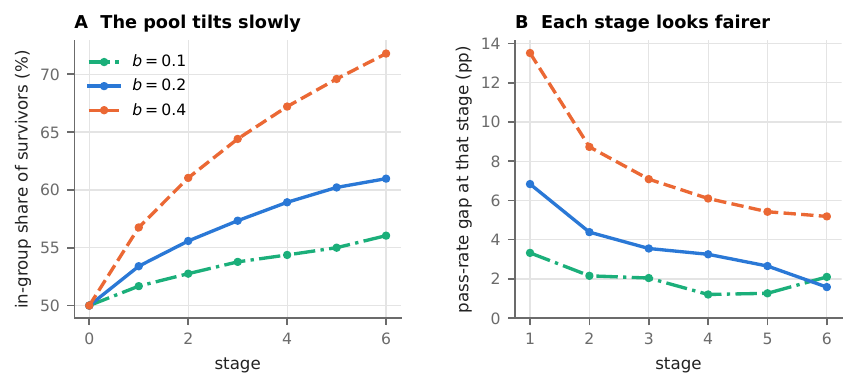}
\caption{\textbf{A fixed series of six biased gates.} A: the in-group's share of survivors by stage for
three per-stage biases $b$. B: the pass-rate gap at each stage, which shrinks as the pool tilts. The
stage-$6$ values for $b=0.1$ carry visible sampling noise.}
\label{fig:static}
\end{figure}

\paragraph{A fixed series of biased gates.} Without feedback, the natural model of a biased pipeline is
a fixed series of \gate{threshold} gates, each carrying the same small bias. Six stages each advance
half the survivors, and every evaluator adds $b$ for the in-group, with $\sigma_\varepsilon=0.6$ and
$\sigma_T=1$. At $b=0.2$ the pass-rate gap, what an auditor could see at one stage, is $6.8$ percentage
points at the first gate and falls to $1.6$ by the sixth, as the surviving pool tilts. The in-group's
share of survivors rises from $50\%$ to $61\%$, and its composition odds after stages $0$ to $6$ are
$1.00$, $1.15$, $1.25$, $1.34$, $1.44$, $1.51$ and $1.56$ (Figure~\ref{fig:static}). A fixed series of
biased gates drifts modestly, and each stage looks fairer than the last; the feedback of
\S\ref{sec:homophily} can sustain a majority for many replacements.

\clearpage
\section{The last decision in full}
\label{app:execfull}

\paragraph{Design.} Each firm draws $4{,}000$ applicants, half of them with the tracked attribute $z=1$,
unrelated to quality. Two panels of eight each average their members' judgments and advance $40\%$ of
the candidates before them. The panels are different people at each stage: every panel draws its
biases afresh, equicorrelated at $\rho_\beta$ within the panel and independent across panels. The final
evaluator then re-scores every survivor and advances $40\%$, ignoring the panels' scores. The
executive's bias is drawn $\mathcal N(0,\sigma_\beta^2)$ independently of the panels, so it is unbiased
on average only across firms. The measure is the spread across firms of the tracked group's share of
hires, in percentage points, and the hires' mean true success probability.

\paragraph{One person's bias and one person's noise.} A single executive differs from a fresh panel of
eight in two things at once: it carries one person's whole bias variance, where the panel's average
carries $[\rho_\beta+(1-\rho_\beta)/8]$ of it, and one person's whole occasion noise, where the panel's
carries an eighth of its variance. Crossing the two on the same survivors separates them
(Table~\ref{tab:cross}). At matched noise, one person's bias adds spread at every $\rho_\beta$. One
person's noise lowers the spread and costs about two points of hire quality. The executive's lower spread after correlated panels therefore
comes from its noise, and it is paid for in hire quality. The same split appears in the validity of a
gate's score, $v_k=\sigma_T/\sqrt{\sigma_T^2+\Var(z)\sigma_\beta^2[\rho_\beta+(1-\rho_\beta)/k]
+\sigma_\varepsilon^2/k}$: at these parameters ($\sigma_T=1$, $\Var(z)=0.25$, $\sigma_\beta=0.5$,
$\sigma_\varepsilon=0.7$, $\rho_\beta=0$), $v_8=0.967$ and $v_1=0.803$, and the gap between them is
mostly noise, $\sigma_\varepsilon^2=0.49$ against a bias contribution of $\Var(z)\sigma_\beta^2=0.0625$.

\begin{table}[h]
\caption{\textbf{The final evaluator's bias crossed with its noise} ($2{,}000$ firms per $\rho_\beta$).
Each cell: spread across firms of the tracked share of hires (percentage points) / the hires' mean true
success probability (\%). Standard errors: $0.10$--$0.31$ points for spreads, below $0.02$ for mean
success.}
\label{tab:cross}
\centering
\small
\begin{tabular}{@{}lcccc@{}}
\toprule
$\rho_\beta$ of panels & \shortstack{eight judges' bias,\\eight judges' noise} & \shortstack{one person's bias,\\eight judges' noise} &
\shortstack{eight judges' bias,\\one person's noise} & \shortstack{one person's bias\\and noise} \\
\midrule
$0.0$ & $8.37$ / $86.21$ & $19.29$ / $85.77$ & $6.53$ / $84.07$ & $13.92$ / $83.93$ \\
$0.2$ & $11.59$ / $86.13$ & $19.01$ / $85.80$ & $9.09$ / $84.03$ & $14.07$ / $83.93$ \\
$0.5$ & $15.46$ / $85.98$ & $18.77$ / $85.81$ & $12.38$ / $83.96$ & $14.64$ / $83.91$ \\
$0.8$ & $17.54$ / $85.88$ & $18.53$ / $85.82$ & $14.49$ / $83.88$ & $15.01$ / $83.87$ \\
\bottomrule
\end{tabular}
\end{table}

\paragraph{Firm performance at the same selection quota.}
Interpret the $n=256$ selections at firm $f$ as equal-stake projects. Let project $j$ pay one
on success and zero otherwise, with success probability $q_j$ given the project qualities
and selections. If $V_f$ is payoff per project and $A_f$ is the selected portfolio, then
\begin{equation}
 \E[V_f\mid A_f,(q_j)]=\bar q_f=\frac{1}{n}\sum_{j\in A_f}q_j .
 \label{eq:portfolio-payoff}
\end{equation}
The crossing in Table~\ref{tab:cross} records $\bar q_f$ separately for each of the $2{,}000$
firms in each arm. At $\rho_\beta=0$, the across-firm standard deviation of $\bar q_f$ is
$0.42$ percentage points with a final panel, $0.75$ when one person sets the preferences at
the same noise level, and $0.54$ when that person also supplies the noise. The final cut acts
on exactly the same survivors in all arms. A decision maker can shift the mix toward a favored
group and thereby change which project qualities clear the cut. At $\rho_\beta=0.8$, the same
three spreads are $0.72$, $0.74$ and $0.68$ points. The direction thus depends on which errors
concentration changes. This calculation measures cross-firm variation in expected performance
at equal stakes. Estimating stock-return volatility would also require a model of realized
payoffs through time.

\paragraph{Sampling and selectivity.} An earlier run of the same design compared the executive with a
panels-only arm and with three controls at matched selectivity, over $600$ firms per $\rho_\beta$
(Table~\ref{tab:exec}). With three cuts that each advance $40\%$, the executive arm hires $256$ per firm
and the panels-only arm, with two cuts, $640$. Binomial sampling alone gives a spread of the tracked
share of $3.1$ and $2.0$ points at a share of one half; removing it leaves about $14.0$ and $6.4$ points,
so the gap between those two arms is not a sampling artefact. The arms also differ in selectivity, which
the controls hold fixed, each adding a third $40\%$ cut: on the second panel's score (C1), by a third
panel of eight (C2), and by a third panel with the executive as one of its eight members (C3). At
$\rho_\beta=0$ the executive raises the spread from $8.0$--$9.0$ points in the matched controls to
$14.3$, and both repairs that keep one person's bias out of the last cut work: cutting on the panels'
score (C1, $9.0$) and seating the executive on a panel (C3, $8.5$). This earlier run recorded only the
spread; the crossing above shows that the executive's lower spread at high $\rho_\beta$ is bought with
hire quality.

\begin{table}[h]
\caption{\textbf{The last decision at matched selectivity} ($600$ firms per $\rho_\beta$). Spread across
firms of the tracked group's share of hires, in percentage points (SE $0.2$--$0.6$). Panels only: two
$40\%$ cuts, $640$ hires; other arms: three cuts, $256$ hires. C1: a third cut on the second panel's
score; C2: a third panel of eight; C3: the executive as one member of a third panel.}
\label{tab:exec}
\centering
\small
\begin{tabular}{@{}lccccc@{}}
\toprule
$\rho_\beta$ of panels & panels only & $+$ executive & C1 & C2 & C3 \\
\midrule
$0.0$ & $6.74$ & $14.34$ & $8.99$ & $8.01$ & $8.45$ \\
$0.2$ & $10.24$ & $14.11$ & $13.33$ & $11.86$ & $10.97$ \\
$0.5$ & $12.71$ & $14.23$ & $16.22$ & $15.31$ & $13.61$ \\
$0.8$ & $15.68$ & $15.47$ & $19.91$ & $17.66$ & $16.04$ \\
\bottomrule
\end{tabular}
\end{table}

\begin{figure}[h]
\centering
\includegraphics[width=0.92\linewidth]{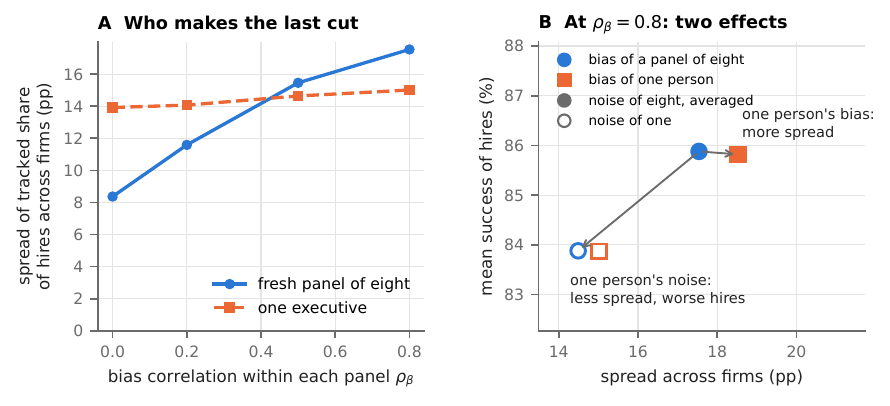}
\caption{\textbf{The last decision.} A: the spread across firms of the tracked group's share of hires,
for a fresh panel of eight and for one executive at the last cut, against the panels' bias correlation.
B: at $\rho_\beta=0.8$, the final evaluator's bias and noise crossed: one person's bias moves the spread
up, and one person's noise moves it down at a cost of about two points in hire quality.}
\label{fig:exec}
\end{figure}

\end{document}